# Heterogeneous SAR–optical fusion for near-real-time land use and land cover mapping under cloud contamination: A novel framework and global benchmark dataset

**Jiangong Xu** [a], **Weibao Xue**[b], **Xiaoyu Yu**[c], **Jun Pan** [a, d, e *], **Xinlian Liang**[a], **Mi Wang**[a, d, e]

[a] *State Key Laboratory of Information Engineering in Surveying, Mapping and Remote Sensing, Wuhan University, Wuhan 430079, China*

[b] *School of Computer Science and Information Engineering, Hefei University of Technology, Hefei 230009, China*

[c] *Hangzhou International Innovation Institute, Beihang University, Hangzhou 311100, China*

[d] *Oriental Space Port Research Institute, Yantai, 265100, China*

[e] *Hubei Luojia Laboratory, Wuhan, 430079, China*

A B S T R A C T

Optical remote sensing imagery is frequently degraded by cloud and cloud-shadow contamination, which limits its reliability for near-real-time land use and land cover (LULC) mapping. Although synthetic aperture radar (SAR) can provide cloud-penetrating structural information, existing SAR–optical fusion methods often assume reliable optical observations and insufficiently address the semantic uncertainty introduced by cloud contamination. To address this issue, we propose CloudLULC-Net, an end-to-end heterogeneous SAR–optical fusion framework that directly predicts LULC maps from cloud-contaminated Sentinel-2 imagery and temporally adjacent Sentinel-1 SAR observations. The proposed network incorporates optical reliability modulation to suppress unreliable optical responses, heterogeneous information adaptive aggregation to model high-order spatial–channel interactions between optical and SAR representations, and a unified semantic mapping transformer to organize fused features in a LULC-oriented latent space. A semantic anchor-guided optimization strategy is further introduced to improve the consistency of intermediate semantic representations. To support this task, we construct CloudLULC-Set, a large-scale benchmark dataset containing 40,223 curated SAR–optical–label triplets with pixel-level LULC annotations across diverse geographic regions and cloud conditions. Experimental results show that CloudLULC-Net achieves an OA of 86.60%, an F1-score of 83.29%, and an mIoU of 73.51%, outperforming representative heterogeneous reconstruction-first and end-to-end SAR–optical mapping methods. Comparisons with existing global LULC products and analyses under different cloud-cover levels further demonstrate the robustness and practical value of CloudLULC-Net for target-date LULC mapping in cloud-prone regions. The project is publicly available at: https://github.com/RSIIPAC/CloudLULC

## 1. Introduction

Accurate and timely land use and land cover (LULC) information is fundamental for monitoring land-surface dynamics and supporting a wide range of Earth observation applications, including agricultural assessment, urban expansion analysis, disaster response, ecological conservation, and natural resource management (Ghorbanian et al., 2020; Wang et al., 2023; Li et al., 2024). Optical remote sensing imagery has long been one of the primary data sources for LULC mapping because it provides rich spectral, spatial, and textural information that is directly related to land-surface properties (Zhu et al., 2017; Ma et al., 2019; Liu et al., 2026). In recent years, a series of regional and global LULC products derived from optical satellite observations, such as ESA WorldCover (Zanaga et al., 2022), Esri Land Cover (Karra et al., 2021), and Google Dynamic World (Brown et al., 2022), have greatly promoted large-scale land cover monitoring. However, optical remote sensing is inherently affected by atmospheric conditions. Clouds and cloud shadows frequently obscure land-surface signals, distort spectral responses, and reduce the availability of valid optical observations, especially in tropical, subtropical, monsoon, and high-latitude regions (Qiu et al., 2019; Williamson et al., 2019). This limitation is particularly critical for near-real-time or target-date LULC mapping, where the generated map is expected to reflect the current or recent land-surface state rather than a temporally delayed or heavily composited representation.

Existing studies have attempted to mitigate cloud-induced observation gaps mainly through several representative paradigms, as summarized in Fig. 1. The first group is homogeneous temporal mapping, which exploits multi-temporal optical observations from the same or similar sensors to recover missing surface information or improve temporal continuity. According to the order in which fusion and classification are performed, these methods can be

* Corresponding author.

*E-mail address:* dd_xjg@whu.edu.cn (J. Xu), xuewb@mail.hfut.edu.cn (W. Xue), yuxiaoyu_buaa@buaa.edu.cn (X. Yu), panjun1215@whu.edu.cn (J. Pan), xinlian.liang@whu.edu.cn (X. Liang), wangmi@whu.edu.cn (M. Wang)

divided into two typical schemes. The first scheme follows a homogeneous fusion-first mapping paradigm, where cloud-contaminated optical images are first reconstructed or temporally fused, and the restored images are then used for LULC classification (Zhu and Woodcock, 2014; Chen et al., 2016; Shu et al., 2025). The second scheme adopts a homogeneous classification-first mapping paradigm, where multiple optical observations are classified independently, and the resulting LULC maps are subsequently integrated to obtain a more complete land cover representation (Li et al., 2024; Zhang et al., 2025b). These homogeneous temporal methods are attractive because they can make full use of existing optical time-series archives without requiring additional sensor modalities. Nevertheless, their effectiveness strongly depends on the availability of sufficient cloud-free or low-cloud optical observations within a suitable temporal window. In persistently cloudy regions, valid optical observations may remain scarce for extended periods. In rapidly changing landscapes, temporally substituted observations may introduce phenological mismatch, seasonal inconsistency, or outdated land-surface information. Moreover, the two-step nature of these methods means that image reconstruction or map-level integration is usually optimized separately from the final classification objective, which may limit the preservation of class-discriminative information for downstream LULC mapping (Xu et al., 2024; Xu et al., 2025).

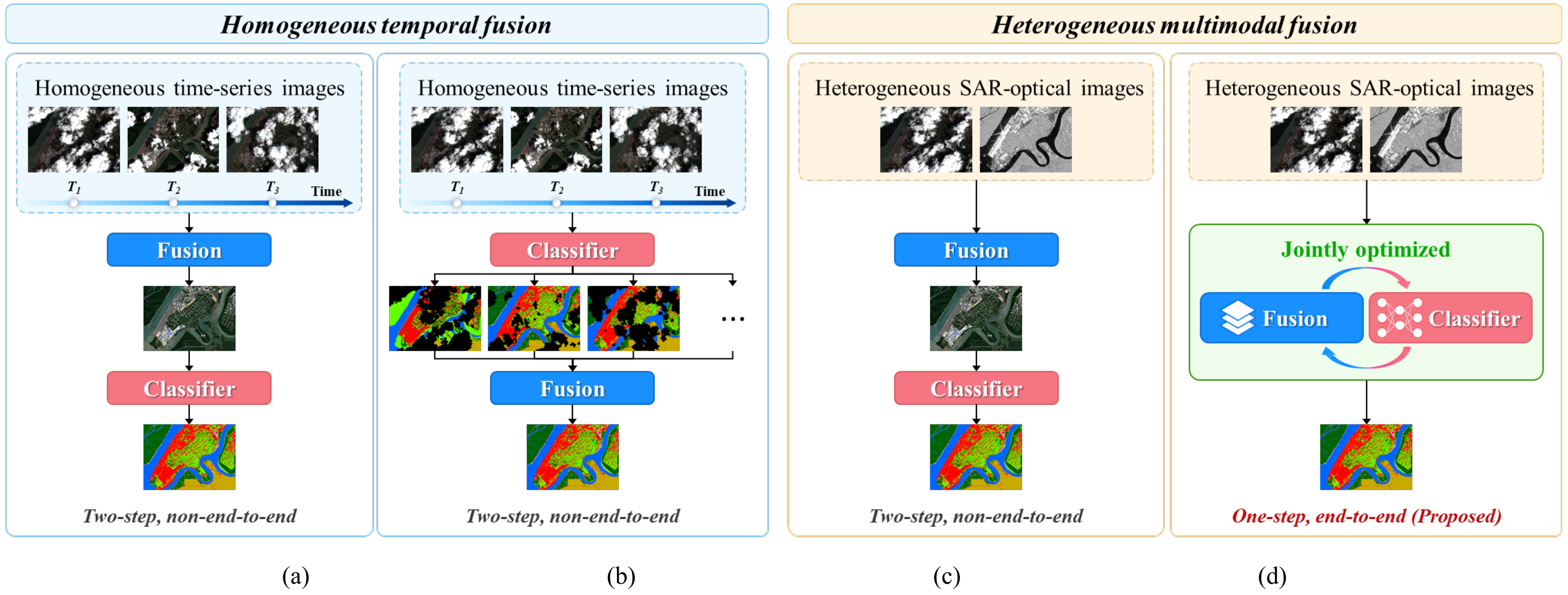

**Fig. 1.** Comparison of representative paradigms for LULC mapping under cloud contamination. (a) Homogeneous fusion-first mapping, where multi-temporal optical observations are first fused or reconstructed and then classified. (b) Homogeneous classification-first mapping, where multi-temporal optical observations are first classified independently and then integrated at the map level. (c) Heterogeneous reconstruction-first mapping, where SAR–optical fusion is used to reconstruct cloud-contaminated optical imagery before LULC classification. (d) Heterogeneous end-to-end mapping, where cloud-contaminated optical imagery and SAR observations are jointly optimized in a unified classification framework.

The second group is heterogeneous SAR–optical mapping, which introduces observations from sensors with different imaging mechanisms to complement cloud-contaminated optical imagery. Synthetic aperture radar (SAR) has attracted increasing attention in this context because it is an active microwave imaging system capable of acquiring land-surface observations under cloudy conditions and independently of solar illumination (Kattenborn et al., 2019; Li et al., 2021). Compared with optical imagery, SAR provides physically different information related to surface structure, roughness, geometry, and moisture conditions (Moreira et al., 2013). Therefore, SAR can provide complementary cues for identifying land cover categories when optical spectral information is partially or completely degraded by clouds. Existing heterogeneous SAR–optical mapping methods can also be broadly divided into two representative schemes. The first scheme is heterogeneous reconstruction-first mapping, where SAR data are first used to assist cloud removal or optical image reconstruction, and LULC classification is then performed on the reconstructed optical images (Li et al., 2023; Pan et al., 2024; Xu et al., 2025; Zhang et al., 2025c). This strategy improves the visual completeness of optical imagery, but the cloud removal process is usually optimized using pixel-wise reconstruction losses and may not preserve high-level semantic information required for downstream classification. The second scheme is heterogeneous end-to-end mapping, where cloud-contaminated optical imagery and SAR imagery are directly fused for LULC classification, allowing the fusion and classification processes to be jointly optimized toward the final mapping objective (Xu et al., 2024; Zhang et al., 2024; Yu et al., 2025; Wei et al., 2026). Compared with reconstruction-first pipelines, this paradigm can reduce the risk of semantic degradation caused by separate reconstruction and classification procedures.

Despite these advances, robust LULC mapping from cloud-contaminated SAR–optical imagery remains challenging. First, SAR and optical imagery are heterogeneous data sources with substantial differences in imaging mechanism, radiometric characteristics, noise properties, geometric distortions, and semantic representation (Wu et al., 2022; Zhang et al., 2025a). These differences may lead to domain gaps and feature misalignment during fusion, especially when the two observations are acquired on different dates or under different imaging geometries. Second, cloud-contaminated optical imagery provides spatially non-uniform information. Cloud-free or weakly contaminated regions still contain valuable spectral and textural cues, whereas cloud-covered regions may provide incomplete or misleading surface information (Xu et al., 2024). Directly fusing optical and SAR features without considering this spatially varying reliability may cause the model to overuse corrupted optical features in cloud-covered areas or underuse informative optical features in clear-sky areas. Third, many existing SAR–optical fusion models are designed under clear-sky or high-quality optical conditions and therefore do not sufficiently address the semantic uncertainty introduced by cloud contamination (Li et al., 2022a; Zhang et al., 2024). Finally, although several SAR–optical or SAR-based benchmark datasets have been released, they mainly focus on building extraction, cloud-free land cover classification, or SAR-only mapping, while public datasets specifically designed for cloud-contaminated SAR–optical LULC mapping remain limited in geographic coverage, cloud-condition diversity, annotation transparency, and standardized evaluation protocols (Shermeyer et al., 2020; Xia et al., 2025).

To address these issues, this study proposes CloudLULC-Net, a heterogeneous SAR–optical fusion framework for near-real-time LULC mapping under cloud contamination. Instead of treating cloud removal and LULC classification as two independent tasks, CloudLULC-Net directly integrates cloud-contaminated optical imagery and temporally adjacent SAR imagery in an end-to-end manner and optimizes the fused representation toward the final LULC mapping objective. The key idea is to exploit the complementary strengths of optical and SAR observations while reducing the negative influence of cloud-induced optical degradation and cross-modal discrepancy. Specifically, the model introduces an optical reliability modulation module to adaptively regulate the contribution of cloud-contaminated optical features, a heterogeneous information adaptive aggregation module to model high-order spatial–channel interactions between SAR and optical representations, and a unified semantic mapping transformer to organize the fused features in a LULC-oriented latent space. In addition, a semantic anchor-guided optimization strategy is designed to regularize the intermediate semantic representation using label-derived semantic anchors. Through this design, CloudLULC-Net can preserve useful optical spectral information in less-contaminated regions, incorporate SAR-derived structural cues in cloud-obscured areas, and improve the semantic consistency of LULC predictions under varying cloud conditions.

In addition to the methodological framework, this study constructs CloudLULC-Set, a large-scale benchmark dataset for cloud-contaminated SAR–optical LULC mapping. The dataset contains temporally paired Sentinel-2 optical imagery, Sentinel-1 SAR imagery, cloud-condition information, and pixel-level LULC annotations across geographically diverse regions. To improve the reliability and reproducibility of the benchmark, the dataset construction procedure considers regional and seasonal diversity, cloud coverage distribution, label harmonization, quality control, and a fixed training/validation/test split. By providing both a dedicated dataset and a standardized evaluation protocol, CloudLULC-Set supports systematic research on heterogeneous SAR–optical fusion for LULC mapping under real-world cloud contamination. Extensive experiments based on this benchmark demonstrate that CloudLULC-Net achieves robust classification performance across different cloud coverage levels, land cover categories, and representative regions, showing its potential for target-date LULC mapping in cloud-prone areas.

The main contributions of this study are summarized as follows:

1) We formulate near-real-time LULC mapping under cloud contamination as a heterogeneous SAR–optical fusion problem and systematically distinguish it from homogeneous temporal fusion and heterogeneous reconstruction-first mapping paradigms.
2) We construct CloudLULC-Set, a large-scale benchmark dataset for cloud-contaminated SAR–optical LULC mapping, which provides temporally paired Sentinel-2/Sentinel-1 observations, pixel-level LULC annotations, cloud-condition statistics, and a standardized evaluation protocol.
3) We propose CloudLULC-Net, an end-to-end heterogeneous SAR–optical fusion framework that directly predicts LULC maps from cloud-contaminated optical imagery and temporally adjacent SAR observations, avoiding the semantic degradation that may occur in reconstruction-first pipelines.
4) We design optical reliability modulation and heterogeneous information adaptive aggregation modules to suppress unreliable optical responses and enhance high-order spatial–channel interactions between SAR and optical representations under cloud-affected conditions.
5) We introduce a unified semantic mapping transformer and semantic anchor-guided optimization strategy to align fused SAR–optical features with a LULC-oriented semantic latent space, improving semantic consistency and classification robustness.

The remainder of this paper is organized as follows. Section 2 reviews related work. Section 3 introduces the construction of CloudLULC-Set. Section 4 presents the proposed heterogeneous SAR–optical fusion framework. Section 5 reports experimental results and analyses. Section 6 discusses the findings and limitations, and Section 7 concludes the paper.

## 2. Related work

### 2.1. SAR–optical fusion for cloud-contaminated remote sensing interpretation

SAR–optical fusion has become an important direction in remote sensing interpretation because the two sensors provide complementary observations of the land surface. Optical imagery records reflected solar radiation and contains abundant spectral, spatial, and textural information, whereas SAR measures microwave backscatter and is sensitive to surface structure, roughness, geometry, and moisture-related properties (Verhoest et al., 2008; Moreira et al., 2013). This complementarity has motivated extensive studies on multi-source land cover classification, semantic segmentation, object detection, and change monitoring (Schmitt et al., 2019; Chen and Bruzzone, 2021; Li et al., 2021).

Early SAR–optical fusion methods mainly relied on handcrafted descriptors, pixel-level concatenation, decision-level integration, or traditional machine learning classifiers. With the development of deep learning, dual-branch convolutional networks have been widely adopted to extract modality-specific representations from optical and SAR images, followed by feature-level fusion for land cover classification or semantic segmentation (Schmitt et al., 2019; Chen and Bruzzone, 2021). To improve the adaptiveness of heterogeneous feature integration, attention-based and gated fusion mechanisms have been introduced to selectively enhance informative modality-specific features and suppress redundant or noisy responses (Li et al., 2021; Li et al., 2022a; Wei et al., 2024). More recently, transformer-based, multi-scale, and multi-level fusion architectures have been developed to model long-range dependencies and hierarchical cross-modal interactions, further improving the representation capability of SAR–optical fusion networks (Liu et al., 2024; Ma et al., 2024b; Zhang et al., 2024; Li et al., 2025b).

These studies demonstrate the effectiveness of SAR–optical complementarity in improving remote sensing interpretation. However, most existing SAR–optical fusion models are developed under clear-sky or high-quality optical conditions, implicitly assuming that optical features are reliable across the entire image. This assumption becomes problematic under cloud contamination, where clear or weakly contaminated regions still provide useful spectral and textural cues, while cloud-covered regions may contain incomplete or misleading information. Directly applying conventional fusion strategies to cloud-contaminated optical inputs may cause the model to overuse degraded optical features in occluded areas or underuse reliable optical cues in unobstructed areas. Therefore, SAR–optical fusion for cloud-contaminated LULC mapping requires a task-oriented design that jointly considers heterogeneous feature integration, spatially varying optical reliability, and the final classification objective.

### 2.2. High-order heterogeneous interaction and semantic latent mapping

Effective SAR–optical fusion requires the model to capture both local spatial details and global contextual dependencies. Conventional convolutional operations are effective in extracting local textures, edges, and structural patterns, but their limited receptive fields make it difficult to explicitly model long-range dependencies. Self-attention and transformer-based architectures address this limitation by computing pairwise relationships among spatial tokens and have been widely adopted in semantic segmentation and multimodal remote sensing fusion (Vaswani et al., 2017; Dosovitskiy et al., 2020). Recent SAR–optical fusion transformers further combine convolutional local feature extraction with transformer-based global modeling, while multilevel multimodal fusion architectures introduce hierarchical cross-modal interaction modules to enhance feature representation for remote sensing semantic segmentation (Liu et al., 2024; Ma et al., 2024b).

Despite their effectiveness, standard self-attention mainly models pairwise token interactions through query–key correlations and often suffers from quadratic computational complexity when processing high-resolution remote sensing images. To go beyond pairwise interaction, high-order interaction mechanisms have recently been explored for multimodal image fusion. These methods aim to iteratively model more complex dependencies among spatial positions, channels, and modalities, and have shown potential in exploiting complementary information from heterogeneous data sources (Zhou et al., 2024). For SAR–optical LULC mapping under cloud contamination, such high-order interaction is particularly meaningful because the relationship between optical spectral cues and SAR structural responses is spatially heterogeneous and strongly affected by cloud-induced degradation. Nevertheless, existing high-order interaction methods are mostly designed for general multimodal image fusion rather than target-oriented SAR–optical LULC classification under cloud-contaminated conditions.

Frequency-domain modeling provides another useful perspective for improving global representation and computational efficiency. Fourier-based operators and spectral transformations have been introduced into deep vision models to enlarge receptive fields and capture global contextual information with relatively low computational cost (Chi et al., 2020; Lee-Thorp et al., 2022; Chen et al., 2024). In multimodal fusion, frequency-aware mechanisms have also been explored to separate modality-shared structural information from modality-specific radiometric characteristics (Zheng et al., 2025). This is relevant to SAR–optical fusion because the two modalities differ substantially in radiometric response but may share geometric or structural patterns. However, most frequency-aware fusion methods focus on generic image fusion, restoration, or frequency decomposition, with limited consideration of joint spatial–channel dependency modeling across heterogeneous SAR–optical representations for semantic LULC mapping.

In addition to feature-level interaction, recent multimodal learning studies have increasingly emphasized the importance of organizing heterogeneous features in a shared semantic latent space. Such latent-space modeling can reduce the gap between modality-specific representations and task-oriented semantic categories, especially when input modalities differ substantially in imaging mechanism or reliability. Anchor-based latent regression has also been explored to stabilize cross-modal representation learning by providing structured targets in the latent space (Chen et al., 2026). However, most existing latent mapping or anchor-guided strategies are designed for modality translation or general multimodal representation learning rather than dense LULC prediction from cloud-contaminated SAR–optical observations. This motivates the unified semantic mapping transformer and semantic anchor-guided optimization strategy proposed in this study, where fused SAR–optical tokens are explicitly organized and regularized in a LULC-oriented semantic latent space.

### 2.3. Benchmark datasets for SAR–optical and cloud-contaminated LULC mapping

Public benchmark datasets are essential for fair comparison, reproducible evaluation, and methodological progress in SAR–optical remote sensing interpretation. Existing datasets related to this topic can be broadly grouped into three categories. The first category includes general SAR–optical or multi-sensor datasets for land cover interpretation, such as WHU-OPT-SAR (Li et al., 2022a), MultiSenGE (Ma et al., 2022), Hunan Dataset (Li et al., 2022b), PIE-RGB-SAR (Zhang et al., 2024), and DDHR-SK (Ren et al., 2022). These datasets provide paired or co-located optical and SAR observations and have supported studies on multi-source land cover classification, semantic segmentation, and cross-modal representation learning. However, most of them are constructed under clear-sky or high-quality optical conditions, rely on region-specific sampling, or use land cover products that may not be temporally matched with the image acquisition dates. Therefore, they are not specifically designed for evaluating near-real-time LULC mapping from real cloud-contaminated optical inputs.

The second category consists of cloud-oriented remote sensing datasets, including optical cloud detection/removal datasets and SAR-assisted optical cloud removal datasets. For example, SEN12MS-CR (Ebel et al., 2020) and M3R-CR (Xu et al., 2023) provide paired cloudy optical, cloud-free optical, and SAR observations for multimodal cloud removal. These datasets are valuable for cloud mask learning, optical image restoration, and SAR-assisted cloud removal. Nevertheless, their primary objective is usually image reconstruction or cloud processing rather than pixel-level LULC mapping. Moreover, when semantic labels are included or compatible with these datasets, they are often derived from annual land cover products, which may introduce temporal inconsistency with the acquisition dates of the cloud-contaminated optical and SAR observations.

The third category includes large-scale SAR–optical alignment or multi-task datasets, such as SOMA-1M (Wu et al., 2026), which provides massive paired SAR–optical images and supports tasks including image matching, image fusion, SAR-assisted cloud removal, and cross-modal translation. These datasets are valuable for developing general multimodal representation models and cross-modal alignment methods, but they are not specifically tailored to near-real-time LULC mapping under real cloud contamination. In particular, they usually do not jointly provide naturally cloud-contaminated optical observations, temporally adjacent SAR images, target-date pixel-level LULC annotations, cloud-condition statistics, and a standardized evaluation protocol.

Therefore, despite the progress made by existing datasets, a dedicated benchmark remains necessary for cloud-contaminated SAR–optical LULC mapping. Such a benchmark should jointly provide naturally cloud-contaminated optical observations, temporally adjacent SAR images, target-date pixel-level LULC annotations, cloud-condition statistics, and standardized benchmark settings. These requirements are essential for systematically assessing whether a model can directly predict reliable LULC maps from degraded optical observations rather than relying on cloud-free optical inputs or annual land cover products.

## 3. Benchmark dataset construction

To support target-date LULC mapping under cloud-contaminated conditions, we constructed CloudLULC-Set, a dedicated SAR–optical benchmark dataset integrating cloud-contaminated Sentinel-2 optical imagery, temporally adjacent Sentinel-1 SAR observations, and pixel-level LULC annotations. Unlike conventional SAR–optical datasets that are mainly designed for cloud-free land cover classification, image matching, or general multimodal fusion, CloudLULC-Set focuses on a more challenging and practical scenario in which optical observations are partially or heavily degraded by clouds and cloud shadows.

Table 1 compares CloudLULC-Set with representative SAR–optical datasets related to cloud-contaminated LULC mapping. Existing datasets have provided valuable resources for SAR-assisted cloud removal, multimodal image fusion, and land cover interpretation. However, most of them are not specifically designed for pixel-level target-date LULC mapping from naturally cloud-contaminated optical imagery and temporally adjacent SAR observations. For example, SEN12MS-CR (Ebel et al., 2020) and M3R-CR (Xu et al., 2023) are mainly developed for SAR-assisted cloud removal. Although they include or can be associated with semantic land cover information, such labels are often derived from annual products and may be temporally inconsistent with the acquisition dates of the cloudy optical and SAR observations. DDHR-SK (Ren et al., 2022) and SOMA-1M-LR (Wu et al., 2026) provide useful resources for multimodal representation learning or cloud-related tasks, but they are not specifically tailored to target-date LULC mapping under real cloud-contaminated conditions. PIE-RGB-SAR (Zhang et al., 2024) provides high-resolution RGB–SAR samples, but its geographic coverage and task setting differ from the cloud-contaminated Sentinel-1/2 LULC mapping scenario considered in this study. In contrast, CloudLULC-Set jointly provides naturally cloud-contaminated optical inputs, temporally adjacent SAR observations, and pixel-level LULC annotations generated with reference to temporally close cloud-free or near-cloud-free optical imagery, thereby better supporting SAR–optical LULC mapping under real-world cloud contamination.

The details of data sources and preprocessing, annotation protocol, and dataset statistics are presented in the following subsections.

### 3.1. Data sources and preprocessing

The satellite data used for CloudLULC-Set were acquired from the Copernicus Sentinel missions operated by the European Space Agency (ESA). Sentinel-1 and Sentinel-2 were selected because they provide open-access, globally available, and frequently revisited observations with complementary imaging characteristics. Sentinel-2 offers rich multispectral information for LULC interpretation, whereas Sentinel-1 provides cloud-insensitive microwave observations that can complement optical imagery when land-surface signals are obscured by clouds and cloud shadows. This combination makes the Sentinel-1/2 constellation suitable for constructing a reproducible SAR–optical benchmark for near-real-time LULC mapping under cloud-contaminated conditions.

**Table 1**

Comparison of CloudLULC-Set with representative publicly accessible SAR–optical datasets related to cloud-contaminated LULC mapping.

| Dataset | Optical image | | SAR image | | Cloud | Region | Coverage ($km^2$) | Width | Images |
|---|---|---|---|---|---|---|---|---|---|
| | Source | $S_R$ | Source | $S_R$ | | | | | |
| SEN12MS-CR (Ebel et al., 2020) | Sentinel-2 | 10m | Sentinel-1 | 10m | √ | Worldwide | 29,491 | 256 | 122,218 |
| DDHR-SK (Ren et al., 2022) | GaoFen-2* | 1m | GaoFen-3 | 1m | √ # | South Korea | 94 | 256 | 6,173 |
| M3R-CR (Xu et al., 2023) | Planet Scope | 3m | Sentinel-1 | 10m | √ | Worldwide | 51,030 | 300 | 63,000 |
| PIE-RGB-SAR (Zhang et al., 2024) | Google Earth* | 0.5m | GaoFen-3 | 3m | √ | China | 2,869 | 256 | 4,865 |
| SOMA-1M-LR (Wu et al., 2026) | Google Earth* | 8m | Sentinel-1 | 10m | √ # | Worldwide | - | 512 | 357,563 |
| CloudLULC-Set (Ours) | Sentinel-2 | 10m | Sentinel-1 | 10m | √ | Worldwide | 578,690 | 512 | 40,223 |

$S_R$ denotes spatial resolution.

* Only RGB optical bands were included.

\# Cloud scenarios were generated using simulation or task-specific construction.

The optical imagery was obtained from the Sentinel-2 A/B constellation. Level-2A surface reflectance products were used, which provide atmospherically corrected multispectral observations without the cirrus band B10. In this study, all 12 available surface reflectance bands were retained, including B1, B2, B3, B4, B5, B6, B7, B8, B8A, B9, B11, and B12. This setting preserves broad spectral information from visible, red-edge, near-infrared, water-vapor, and shortwave-infrared regions, supporting the discrimination of major land cover types such as vegetation, water, bare surfaces, and built-up land. To ensure pixel-level consistency among optical imagery, SAR observations, and LULC annotations, all selected bands were resampled and aligned to a unified 10 m reference grid during preprocessing.

The SAR imagery was obtained from Sentinel-1 Ground Range Detected (GRD) products acquired in Interferometric Wide (IW) swath mode. Dual-polarization VV and VH observations were used because they provide complementary backscattering information related to land surface structure, roughness, geometry, and moisture conditions. The Sentinel-1 data were processed following standard ESA-recommended procedures, including precise orbit correction, thermal noise removal, radiometric calibration, terrain correction, conversion to backscattering coefficients, logarithmic scaling, and normalization. The processed VV and VH channels were then reprojected and resampled to the same 10 m reference grid as the Sentinel-2 imagery.

For each dataset sample, a cloud-contaminated Sentinel-2 image was paired with a temporally adjacent Sentinel-1 SAR observation. In addition, the nearest available cloud-free Sentinel-2 image was collected as an auxiliary reference for manual annotation and quality inspection, rather than being used as an input to the primary benchmark task. The acquisition dates and temporal intervals among the cloud-contaminated optical image, SAR image, and reference optical image were recorded for each sample to improve the transparency of dataset construction and to reduce uncertainty caused by large temporal gaps.

**3.2. Annotation protocol and label harmonization**

Reliable pixel-level LULC annotations are essential for constructing a benchmark dataset for cloud-contaminated SAR–optical mapping. In CloudLULC-Set, the LULC labels were generated through expert interpretation rather than directly inherited from annual land cover products. This strategy was adopted to reduce the temporal inconsistency that may occur when annual products are used as labels for near-real-time observations. Specifically, temporally adjacent cloud-free Sentinel-2 images were used as the primary reference for manual annotation, while high-resolution optical imagery, Sentinel-1 SAR observations, and existing land cover products were used as auxiliary references for cross-checking and quality control. The reference optical images were selected as close as possible to the acquisition dates of the corresponding cloud-contaminated Sentinel-2 images, thereby reducing label uncertainty caused by seasonal variation or short-term land surface changes.

To ensure semantic consistency across different geographic regions, a unified LULC taxonomy was adopted. The category design was harmonized with widely used land cover classification schemes in global LULC products and remote sensing benchmarks, such as ESA WorldCover (Zanaga et al., 2022), Dynamic World (Brown et al., 2022), Esri Land Cover (Karra et al., 2021), and DeepGlobe (Demir et al., 2018). Considering the semantic distinguishability of Sentinel-1/2 observations and the need for cross-region annotation consistency, thematically similar or easily confused categories were merged into six major land cover classes:

water, forest, grassland, barren land, built-up land, and cropland. Water includes rivers, lakes, reservoirs, ponds, and other visible water bodies. Forest refers to tree-covered areas, including natural forests and plantations. Grassland includes herbaceous vegetation-dominated areas. Cropland refers to cultivated agricultural land, including paddy fields, dryland crops, and other managed agricultural parcels. Barren land includes bare soil, exposed rock, sand, and sparsely vegetated surfaces. Built-up land includes buildings, roads, industrial areas, and other impervious surfaces.

The annotation process was conducted on the unified 10 m reference grid. For each ROI, interpreters first examined the temporally closest cloud-free or near-cloud-free Sentinel-2 imagery to delineate major LULC boundaries. Sentinel-1 SAR imagery was then used as complementary evidence, especially for identifying water bodies, built-up structures, and areas where optical interpretation was affected by residual clouds, shadows, or spectral confusion. Existing land cover products were used only as auxiliary references to support semantic consistency and cross-checking, but they were not directly adopted as ground-truth labels. This procedure helps ensure that the generated labels are more temporally consistent with the cloud-contaminated optical and SAR observations, while avoiding the direct transfer of annual land cover products as ground truth.

### 3.3. Dataset organization and statistics

Selecting geographically representative samples is essential for developing LULC mapping methods with strong generalization ability under diverse cloud-contaminated observation conditions. CloudLULC-Set contains 48 independent regions of interest (ROIs) extracted from Sentinel-1/2 imagery. Each ROI covers approximately $10980\times10980$ pixels, corresponding to a ground area of about $109.8\times109.8$ km$^2$ at a unified 10 m spatial resolution. The total spatial coverage of the dataset is approximately 578,690 km$^2$. As shown in Fig. 2, the selected ROIs are distributed across multiple continents and climatic zones, providing diverse geographic, seasonal, and land cover conditions. Five representative study areas, including the Biobío Region in Chile, Maryland in the United States, Tamil Nadu in India, and Anhui and Jiangxi provinces in China, were further selected to illustrate regional diversity and support subsequent case analyses. The seasonal distribution of the ROIs is also shown in Fig. 2, indicating that the dataset covers different phenological stages and observation conditions.

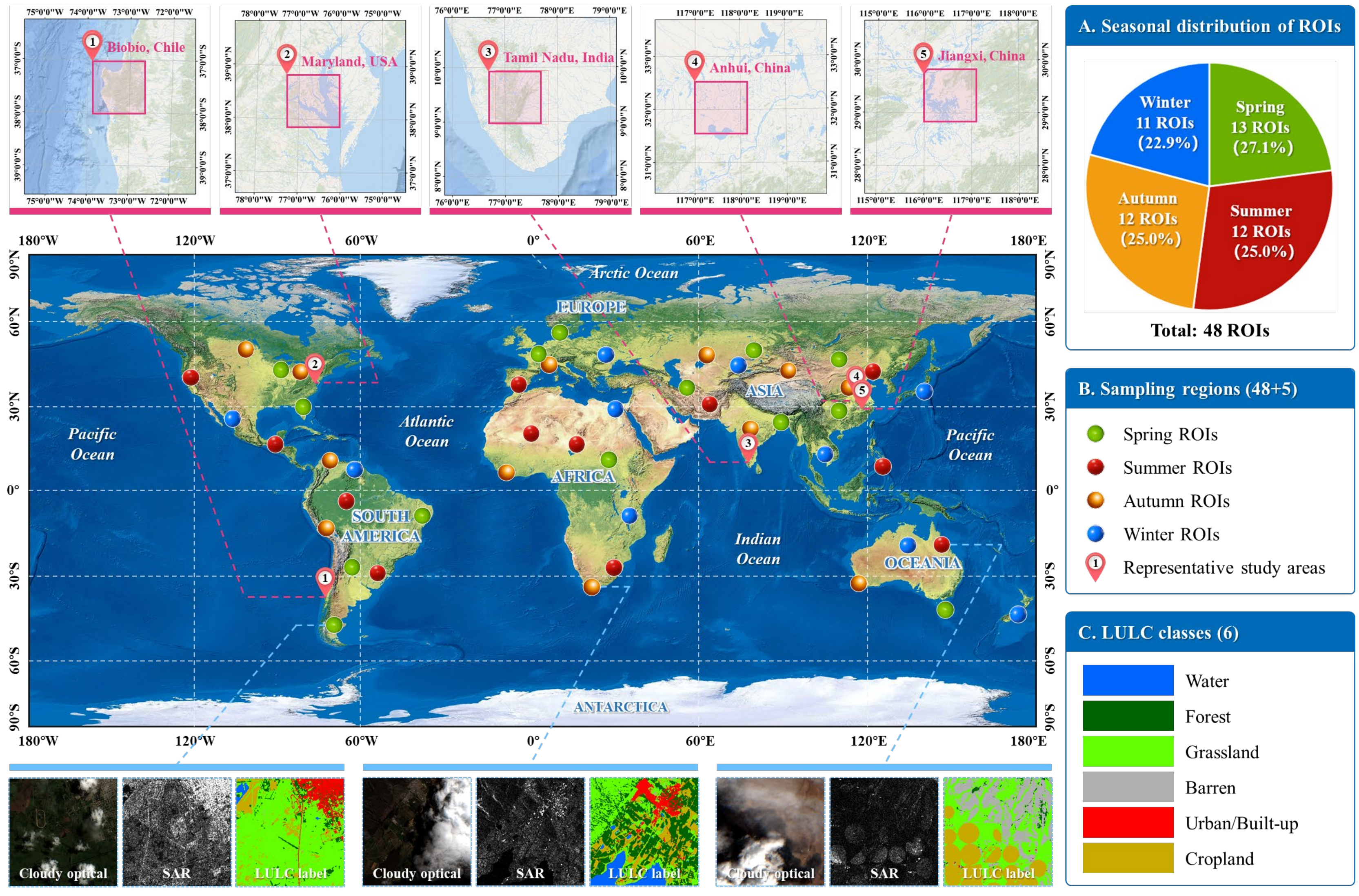

**Fig. 2.** Global spatial distribution, seasonal composition, and representative samples of CloudLULC-Set. The main map shows the globally distributed ROIs across different continents and climatic zones, while the upper insets present five representative study areas, including the Biobío Region, Maryland, Tamil Nadu, Anhui, and Jiangxi. The right panel summarizes the seasonal distribution of ROIs and the legends for LULC classes and sampling seasons. Example triplets of cloud-contaminated optical imagery, SAR imagery, and corresponding LULC labels are shown at the bottom.

Cloud coverage statistics were estimated to characterize the observation difficulty of representative cloud-prone regions. Monthly cloud coverage was calculated for the five representative study areas using Google Earth Engine from October 1, 2023 to September 30, 2024. As shown in Fig. 3, cloud coverage exhibits substantial temporal and regional variability, and the monthly average cloud fraction remains above 40% in many cases. These statistics indicate that cloud contamination is a persistent constraint for optical satellite observations and further justify the need for SAR-assisted LULC mapping under target-date conditions.

In addition to cloud coverage, temporal consistency among heterogeneous observations is another important factor for constructing reliable LULC labels. Large temporal gaps between the cloud-contaminated optical image, the paired SAR observation, and the reference optical image may introduce land surface changes, phenological differences, or annotation uncertainty. Therefore, the acquisition dates of all paired observations were recorded and the temporal intervals were statistically analyzed. The average temporal interval between the cloud-contaminated optical imagery and the corresponding reference optical imagery is 5.17 days, while the mean temporal offset between the optical imagery and the associated SAR observation is 3.35 days. These relatively short intervals help reduce the uncertainty caused by land cover changes between heterogeneous observations and annotation references.

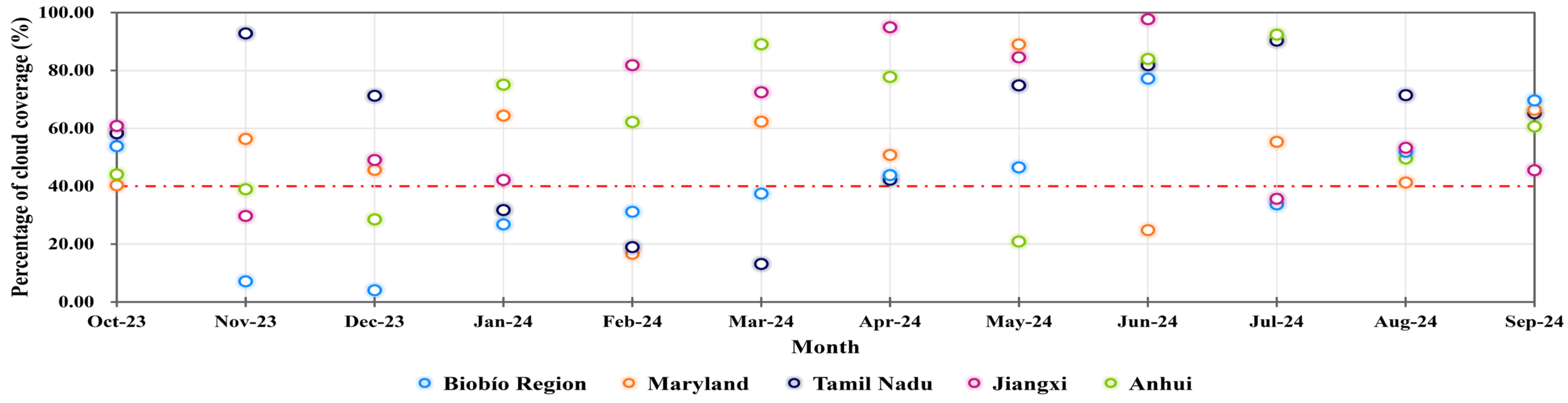

**Fig. 3.** Monthly average cloud coverage of Sentinel-2 imagery across the five representative study areas from October 2023 to September 2024. The statistics were derived using Google Earth Engine and illustrate the temporal and regional variability of cloud contamination.

The class distribution of CloudLULC-Set is summarized in Table 2. The dataset contains six major LULC categories, including forest, grassland, cropland, built-up land, barren land, and water. Forest and grassland account for the largest proportions, with 29.60% and 28.53% of the annotated pixels, respectively, followed by cropland, water, barren land, and built-up land. The built-up category accounts for only 2.52%, reflecting the natural class imbalance of real-world LULC mapping scenarios. Instead of artificially balancing the category distribution, this imbalance was retained to preserve realistic mapping conditions. Accordingly, class-aware metrics such as per-class IoU and F1-score are used together with OA and mIoU in the benchmark evaluation.

**Table 2**
Pixel-level class distribution of LULC annotations in CloudLULC-Set.

| | Forest | Grassland | Cropland | Urban/Built-up | Barren | Water |
|---|---|---|---|---|---|---|
| Percentage | 29.60% | 28.53% | 15.98% | 2.52% | 8.41% | 14.94% |

## 4. Proposed Method

This section presents the proposed CloudLULC-Net for near-real-time LULC mapping from cloud-contaminated SAR–optical observations. CloudLULC-Net directly learns a task-oriented mapping from cloud-contaminated optical imagery and temporally adjacent SAR observations to pixel-level LULC labels. As illustrated in Fig. 4, the main prediction pathway consists of modality-specific feature encoding, optical reliability modulation (ORM), heterogeneous information adaptive aggregation (HIAA), unified semantic mapping transformer (USMT), and LULC prediction decoding. During training, a semantic anchor-guided optimization strategy is further introduced to regularize the intermediate semantic latent representation. The following subsections describe these components in detail.

### 4.1. Overall framework

Given a cloud-contaminated optical image $\mathbf{X}^o \in \mathbb{R}^{H\times W\times C_o}$ and a co-registered SAR image $\mathbf{X}^s \in \mathbb{R}^{H\times W\times C_s}$, the objective of CloudLULC-Net is to predict a pixel-level LULC map $\hat{\mathbf{Y}} \in \{1,\cdots,K\}^{H\times W}$, where $H$ and $W$ denote the spatial dimensions, $C_o$ and $C_s$ denote the numbers of optical and SAR channels, respectively, and $K$ denotes the number of land cover categories. In this study, $C_o = 12$ for the retained Sentinel-2 Level-2A surface reflectance bands, $C_s = 2$ for the Sentinel-1 VV and VH polarizations, and $K = 6$ for the unified LULC taxonomy defined in Section 3.2. The network first produces a class-probability map $\mathbf{P} \in \mathbb{R}^{H\times W\times K}$, and the final LULC map $\hat{\mathbf{Y}}$ is obtained by assigning each pixel to the class with the highest probability.

As shown in Fig. 4, CloudLULC-Net follows an anchor-guided heterogeneous semantic mapping framework. The main inference pathway contains four stages: modality-specific feature encoding, reliability-aware heterogeneous aggregation, unified semantic mapping, and dense LULC prediction. First, the cloud-contaminated optical image and the SAR image are fed into two modality-specific encoders to extract optical and SAR feature representations:

$$\mathbf{F}^o = E_o(\mathbf{X}^o) \tag{1}$$

$$\mathbf{F}^s = E_s(\mathbf{X}^s) \tag{2}$$

where $E_o(\cdot)$ and $E_s(\cdot)$ denote the optical and SAR encoders, respectively. The dual-branch design is adopted because optical and SAR imagery are physically heterogeneous observations. Optical imagery provides rich spectral, spatial, and textural information, whereas SAR imagery provides structural and backscattering information that is less affected by cloud contamination. Therefore, modality-specific feature encoding is used before cross-modal aggregation to preserve the complementary characteristics of the two data sources.

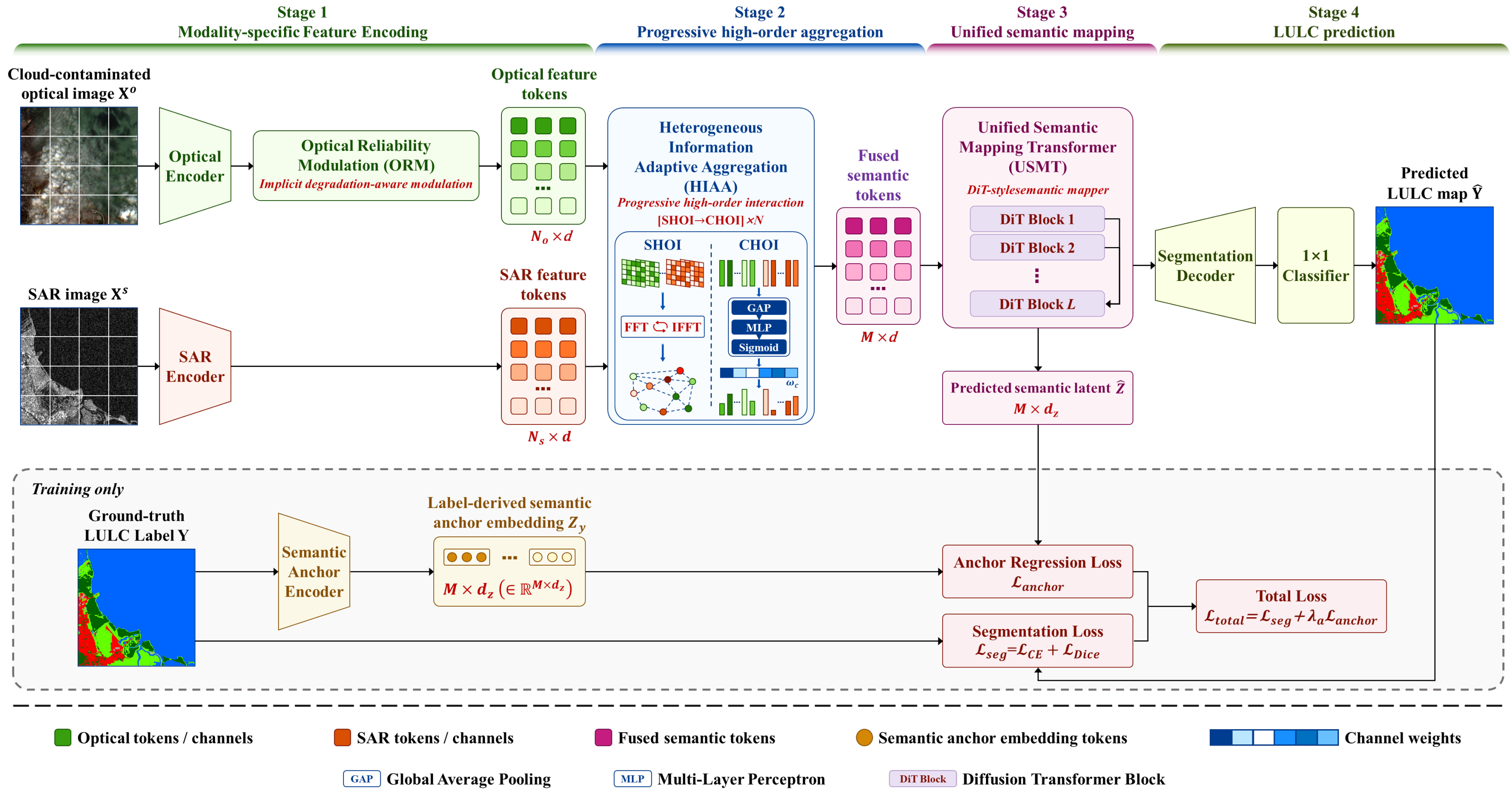

**Fig. 4.** Overall framework of the proposed CloudLULC-Net for cloud- contaminated SAR–optical LULC mapping.

For the optical branch, the extracted optical feature $\mathbf{F}^o$ is further processed by the ORM module to obtain reliability-aware optical features:

$$\tilde{\mathbf{F}}^o = \mathrm{ORM}(\mathbf{F}^o) \tag{3}$$

The ORM module is designed to adaptively regulate optical feature responses under cloud-contaminated conditions. It does not require externally provided cloud or shadow masks, nor does it perform explicit cloud detection. Instead, it learns degradation-aware modulation from optical representations to reduce unreliable responses in cloud-affected regions while retaining useful spectral and textural cues in less-contaminated areas.

The reliability-aware optical features $\tilde{\mathbf{F}}^o$ and the SAR features $\mathbf{F}^s$ are then converted into modality-specific feature tokens and fed into the HIAA module. HIAA is implemented as a stack of $N$ heterogeneous interaction blocks, each consisting of a Spatial High-Order Interaction (SHOI) module followed by a Channel High-Order Interaction (CHOI) module. SHOI models cross-modal spatial dependencies between optical and SAR representations, while CHOI recalibrates spectral–structural channel responses after spatial interaction. Through progressive SHOI–CHOI interaction, HIAA refines heterogeneous SAR–optical representations and produces fused semantic tokens:

$$\mathbf{T}^f = HIAA(\tilde{\mathbf{F}}^o, \mathbf{F}^s), \qquad \mathbf{T}^f \in \mathbb{R}^{M \times d} \tag{4}$$

where $M$ denotes the number of fused semantic tokens and $d$ denotes the token embedding dimension.

After heterogeneous aggregation, the fused semantic tokens are fed into the USMT module. The purpose of USMT is to transform the aggregated SAR–optical tokens into a task-oriented LULC semantic latent space, rather than directly reconstructing cloud-free optical imagery or any target-modality image. This design separates heterogeneous feature aggregation from semantic latent mapping and helps organize fused representations before dense prediction:

$$\hat{\mathbf{Z}} = \mathrm{USMT}(\mathbf{T}^f) \tag{5}$$

where $\hat{\mathbf{Z}} \in \mathbb{R}^{M \times d_z}$ denotes the predicted semantic latent representation and $d_z$ denotes the latent embedding dimension.

Finally, the predicted semantic latent representation $\hat{\mathbf{Z}}$ is passed through a segmentation decoder and a $1 \times 1$ classifier to generate the class-probability map:

$$\mathbf{P} = Softmax\left(\mathrm{Cls}\left(\mathrm{Dec}(\hat{\mathbf{Z}})\right)\right) \tag{6}$$

where $\mathrm{Dec}(\cdot)$ denotes the segmentation decoder, $\mathrm{Cls}(\cdot)$ denotes the $1 \times 1$ classifier, and $\mathbf{P} \in \mathbb{R}^{H \times W \times K}$. The final LULC map is obtained by:

$$\hat{\mathbf{Y}}(m,n) = \arg\max_k \mathbf{P}(m,n,k) \tag{7}$$

During training, $\mathbf{P}$ is supervised by pixel-level LULC annotations, while $\hat{\mathbf{Z}}$ is further constrained by a label-derived semantic anchor. The detailed designs of ORM, HIAA, USMT, and semantic anchor-guided optimization are introduced in the following subsections.

### 4.2. Optical reliability modulation

Cloud-contaminated optical imagery provides spatially non-uniform information for LULC mapping. In cloud-free or weakly contaminated regions, optical features can still provide discriminative spectral, textural, and vegetation-related cues. In contrast, cloud-covered, shadow-affected, or haze-contaminated regions may contain distorted or incomplete optical responses, which can mislead feature fusion and semantic prediction. Directly fusing such optical features with SAR features may cause the network to overuse unreliable optical responses in degraded areas or underuse informative optical cues in less-contaminated regions. To mitigate this problem, we introduce an optical reliability modulation (ORM) module before heterogeneous SAR–optical aggregation.

Given the optical feature $\mathbf{F}^o$ extracted by the optical encoder, ORM learns a feature-level reliability response from the optical representation itself. Instead of explicitly detecting clouds or shadows, ORM estimates a latent reliability gate through a lightweight modulation branch:

$$\mathbf{R}^o = \sigma(\Phi_r(\mathbf{F}^o)) \quad (8)$$

where $\Phi_r(\cdot)$ denotes a learnable reliability estimation function implemented by convolutional projections, $\sigma(\cdot)$ denotes the sigmoid activation, and $\mathbf{R}^o$ represents the learned optical reliability response. The values of $\mathbf{R}^o$ are constrained to the range [0, 1], where larger values indicate higher feature reliability.

To avoid simply discarding degraded optical information, ORM further introduces a residual contextual branch to provide an alternative transformed representation of the optical feature:

$$\mathbf{C}^o = \Phi_c(\mathbf{F}^o) \quad (9)$$

where $\Phi_c(\cdot)$ ddenotes a residual context transformation implemented by convolutional operations. The reliability-aware optical feature is then obtained by adaptively balancing the original optical response and the contextual residual response:

$$\tilde{\mathbf{F}}^o = \mathbf{R}^o \odot \mathbf{F}^o + (1 - \mathbf{R}^o) \odot \mathbf{C}^o \quad (10)$$

where $\odot$ denotes element-wise multiplication, and $\tilde{\mathbf{F}}^o$ is the modulated optical feature. In this formulation, $\mathbf{R}^o$ controls the contribution of the original optical representation, while $1 - \mathbf{R}^o$ adaptively introduces contextual residual cues when the original optical response is less reliable. Therefore, reliable optical information can be preserved in clear or weakly contaminated regions, whereas degraded responses can be partially replaced by transformed contextual representations in cloud-affected regions.

The modulated optical feature $\tilde{\mathbf{F}}^o$ is then used as the optical input to the HIAA module together with the SAR feature$\mathbf{F}^s$. Since SAR observations are less affected by atmospheric cloud contamination and provide complementary structural and backscattering information, ORM helps establish a more balanced SAR–optical fusion process. Reliable optical cues are retained where available, while the influence of unreliable optical responses is reduced before progressive high-order heterogeneous interaction. In this way, ORM serves as a preparatory reliability regulation step that improves the quality of optical representations for subsequent SAR–optical feature aggregation.

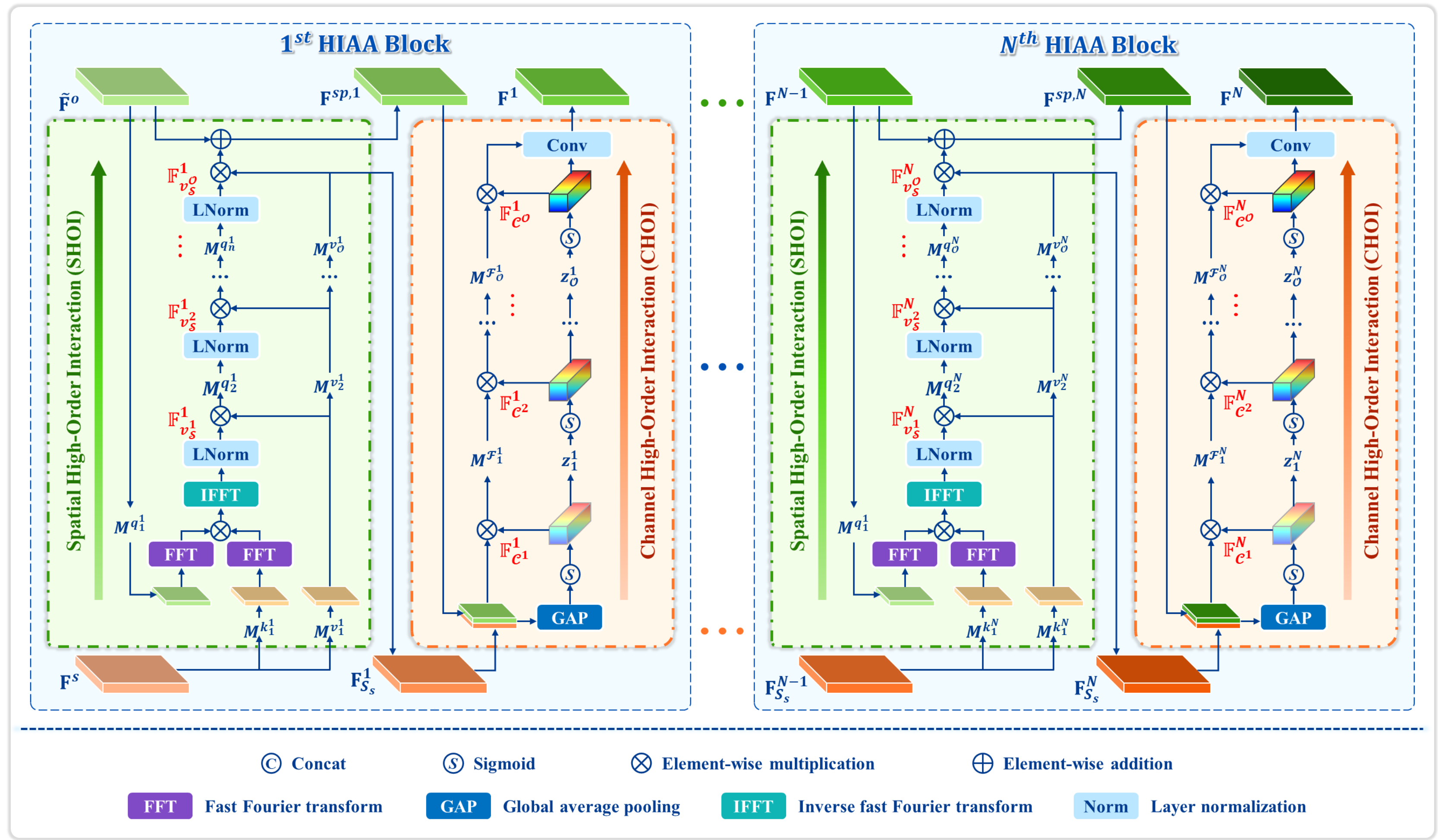

**Fig. 5.** Architecture of the Heterogeneous Information Adaptive Aggregation (HIAA) block, comprising Spatial Higher-Order Interaction (SHOI) and Channel Higher-Order Interaction (CHOI) modules for cross-modal feature integration in both spatial and channel dimensions.

### 4.3. Heterogeneous information adaptive aggregation

After optical reliability modulation, the reliability-aware optical feature $\tilde{\mathbf{F}}^o$ and the SAR feature $\mathbf{F}^s$ are fed into the heterogeneous information adaptive aggregation (HIAA) module. The purpose of HIAA is to progressively integrate heterogeneous SAR–optical information before unified semantic mapping. Different from simple feature concatenation or single-stage attention-based fusion, HIAA first models cross-modal spatial dependencies between the two modalities and then recalibrates channel-wise spectral–structural responses.

As shown in Fig. 5, HIAA is implemented as a stack of (N) HIAA blocks. Each block consists of a Spatial High-Order Interaction (SHOI) module followed by a Channel High-Order Interaction (CHOI) module. Within each block, optical-oriented and SAR-oriented intermediate representations are updated through bidirectional spatial interaction, and the resulting heterogeneous features are further refined through channel-wise recalibration. Through stacked SHOI–CHOI interaction, HIAA progressively refines SAR–optical representations and produces fused semantic tokens for the subsequent unified semantic mapping transformer.

#### 4.3.1. Spatial High-Order Interaction

Remote sensing LULC mapping requires both local spatial details and long-range contextual information. Although convolutional operations are effective in extracting local structures, they have limited ability to explicitly model long-range dependencies. Self-attention can capture global relationships by computing pairwise correlations among spatial tokens, but its quadratic complexity becomes expensive for high-resolution remote sensing features. Moreover, conventional pairwise attention mainly models second-order interactions and may be insufficient for heterogeneous SAR–optical fusion, where cloud-contaminated optical features and SAR structural features exhibit substantial modality discrepancies. Recent multimodal fusion studies have also shown that exploring high-order interactions in spatial and channel dimensions can better exploit cross-modality complementarity than conventional pairwise attention (Zheng et al., 2024).

To address this issue, the SHOI module is designed to model cross-modal spatial dependencies in a recursive manner. Given the optical and SAR features entering the $i$-th HIAA block, denoted as $\mathbf{F}^{o,i-1}$ and $\mathbf{F}^{s,i-1}$, SHOI first projects them into query-, key-, and value-like representations through learnable transformations. Instead of directly computing dense dot-product attention in the spatial domain, SHOI introduces Fourier-transform-assisted correlation modeling to efficiently capture long-range spatial interactions. Frequency-domain modeling has recently been investigated in transformer-based image restoration and fusion to enhance global dependency modeling through efficient spectral operations (Shang et al., 2024). A simplified optical-oriented cross-modal spatial correlation can be expressed as:

$$\boldsymbol{C}^{sp,i} = \mathfrak{F}^{-1}\left(\mathfrak{F}\left(\mathbf{F}^{o,i-1}\mathcal{M}_q^i\right)\otimes\overline{\mathfrak{F}\left(\mathbf{F}^{s,i-1}\mathcal{M}_k^i\right)}\right) \tag{11}$$

where $\mathfrak{F}(\cdot)$ and $\mathfrak{F}^{-1}(\cdot)$ denote the Fast Fourier Transform (FFT) and inverse Fast Fourier Transform (IFFT), respectively, $\overline{(\cdot)}$ denotes complex conjugation, $\otimes$ denotes element-wise multiplication in the frequency domain, and $\mathcal{M}_q^i$ and $\mathcal{M}_k^i$ are learnable projection matrices in the $i$-th HIAA block. The inverse transform yields a real-valued spatial correlation response, which is used to estimate long-range cross-modal dependencies without explicitly constructing a dense spatial attention matrix.

The obtained correlation response is normalized and used to modulate the value representation from the complementary modality:

$$\mathbf{F}^{o,sp,1,i} = \mathrm{LNorm}\left(\boldsymbol{C}^{sp,i}\right)\odot\left(\mathbf{F}^{s,i-1}\mathcal{M}_v^i\right) \tag{12}$$

where $\mathrm{LNorm}(\cdot)$ denotes layer normalization, $\mathcal{M}_v^i$ is a learnable value projection matrix, and $\odot$ denotes element-wise multiplication. The output $\mathbf{F}^{o,sp,1,i}$ represents the initial optical-oriented spatial interaction feature, in which SAR-derived structural cues are introduced through cross-modal spatial correlation.

Similarly, the SAR-oriented spatial interaction feature can be obtained by reversing the roles of optical and SAR features:

$$\mathbf{F}^{s,sp,1,i} = \mathrm{SHOI}_s\left(\mathbf{F}^{s,i-1},\mathbf{F}^{o,i-1}\right) \tag{13}$$

where $SHOI_s(\cdot)$ denotes the SAR-oriented spatial interaction operation. This bidirectional formulation enables the spatial interaction process to update both optical-oriented and SAR-oriented representations, rather than only enhancing one modality.

To go beyond a single pairwise interaction, SHOI applies the above spatial interaction recursively within each block. For the $\sigma$-th order interaction, where $2 \le \sigma \le O$, the recursive spatial interaction can be generally written as:

$$\mathbf{F}^{o,sp,\sigma,i} = \Psi_o^{\sigma}\left(\mathbf{F}^{o,sp,\sigma-1,i},\mathbf{F}^{s,sp,\sigma-1,i}\right) \tag{14}$$

$$\mathbf{F}^{s,sp,\sigma,i} = \Psi_s^{\sigma}\left(\mathbf{F}^{s,sp,\sigma-1,i},\mathbf{F}^{o,sp,\sigma-1,i}\right) \tag{15}$$

where $\Psi_o^{\sigma}(\cdot)$ and $\Psi_s^{\sigma}(\cdot)$ denote the $\sigma$-th recursive spatial interaction functions for the optical-oriented and SAR-oriented branches, respectively, and $O$ denotes the maximum interaction order. The final outputs of SHOI in the $i$-th HIAA block are defined as:

$$\mathbf{F}^{o,sp,i} = \mathbf{F}^{o,sp,O,i} \tag{16}$$

$$\mathbf{F}^{s,sp,i} = \mathbf{F}^{s,sp,O,i} \tag{17}$$

Through this recursive design, SHOI progressively captures higher-order spatial dependencies between SAR and optical representations. This design facilitates the exchange of SAR structural cues and optical spatial details, thereby improving spatial representation under cloud-contaminated conditions.

4.3.2. Channel High-Order Interaction

After spatial high-order interaction, the spatially updated optical-oriented and SAR-oriented features $\mathbf{F}^{o,sp,i}$ and $\mathbf{F}^{s,sp,i}$ are further processed by the CHOI module. While SHOI focuses on spatial dependency modeling, CHOI aims to recalibrate channel-wise responses by modeling dependencies between optical spectral cues and SAR structural backscattering cues. This design is also consistent with recent multimodal fusion studies that emphasize the importance of jointly exploiting spatial and channel interactions for cross-modality information aggregation (Ma et al., 2024b; Zhang et al., 2024). This is important because different feature channels may contribute unequally to LULC discrimination under cloud-contaminated conditions. For example, optical-related channels may be highly informative in cloud-free or weakly contaminated regions, whereas SAR-related channels may provide more reliable structural evidence when optical responses are degraded. In this study, CHOI is designed as a channel-wise high-order recalibration step because it operates on the spatially interacted heterogeneous representations produced by SHOI and is applied progressively across stacked HIAA blocks.

Given the spatially interacted features from SHOI, CHOI first concatenates them along the channel dimension:

$$\mathbf{F}^{c,i} = \mathrm{Concat}\left(\mathbf{F}^{o,sp,i},\mathbf{F}^{s,sp,i}\right) \tag{18}$$

where $\mathrm{Concat}(\cdot)$ denotes channel-wise concatenation, and $\mathbf{F}^{c,i}$ denotes the heterogeneous feature used for channel interaction. It should be noted that this concatenation is performed inside CHOI after spatial interaction, rather than before the HIAA blocks. Therefore, CHOI recalibrates spatially interacted heterogeneous representations rather than simply weighting initially concatenated SAR–optical features.

A global channel descriptor is then obtained by global average pooling:

$$\mathbf{z}^{c,i} = \mathrm{GAP}\left(\mathbf{F}^{c,i}\right) \tag{19}$$

where $\mathrm{GAP}(\cdot)$ denotes global average pooling. The descriptor $\mathbf{z}^{c,i}$ summarizes the global channel statistics of the spatially interacted heterogeneous feature. To model nonlinear channel dependencies, $\mathbf{z}^{c,i}$ is passed through a lightweight multi-layer perceptron and a sigmoid activation to generate channel-wise recalibration weights:

$$\mathbf{w}^{c,i} = \sigma\left(\mathrm{MLP}\left(\mathbf{z}^{c,i}\right)\right) \tag{20}$$

where $\sigma(\cdot)$ denotes the sigmoid activation, $\mathrm{MLP}(\cdot)$ denotes a lightweight channel transformation, and $\mathbf{w}^{c,i}$ represents the learned channel-wise recalibration weights. The weights $\mathbf{w}^{c,i}$ are broadcast along the spatial dimensions and used to recalibrate the heterogeneous feature channels. These weights are not modality-specific masks; rather, they are heterogeneous channel weights learned from the joint optical–SAR representation after spatial interaction.

The channel-recalibrated feature is then obtained as:

$$\mathbf{F}^{out,i} = \mathbf{w}^{c,i} \odot \mathbf{F}^{c,i} \tag{21}$$

where $\odot$ denotes element-wise multiplication. Through this operation, CHOI enhances discriminative spectral–structural channels and suppresses less informative or redundant responses. Compared with directly applying a standard SE block to a single modality, CHOI performs channel recalibration on the spatially interacted heterogeneous feature, thereby allowing optical spectral information and SAR structural information to be adaptively balanced after cross-modal spatial interaction.

By sequentially applying SHOI and CHOI in each HIAA block, the model first establishes cross-modal spatial dependencies and then recalibrates channel-wise heterogeneous responses. For intermediate HIAA blocks, the channel-recalibrated heterogeneous feature $\mathbf{F}^{out,i}$ is projected to update the optical-oriented and SAR-oriented representations for the next block:

$$\left[\mathbf{F}^{o,i}, \mathbf{F}^{s,i}\right] = \Gamma\left(\mathbf{F}^{out,i}\right) \tag{22}$$

where $\Gamma(\cdot)$ denotes a lightweight projection function used to generate the inputs of the next HIAA block. For the final HIAA block, the output feature is transformed into fused semantic tokens:

$$\mathbf{T}^{f} = \Gamma_{T}(\mathbf{F}^{out,N}) \tag{23}$$

where $\Gamma_T(\cdot)$ denotes the token projection function, and $\mathbf{T}^f$ represents the fused semantic tokens fed into the unified semantic mapping transformer. Through $N$ stacked HIAA blocks, the heterogeneous SAR–optical features are progressively refined and transformed into fused semantic tokens for task-oriented LULC semantic representation learning.

**4.4. Unified semantic mapping transformer**

After HIAA, the fused SAR–optical representation contains both optical spectral–textural cues and SAR structural backscattering cues. However, these features are still derived from heterogeneous observations with different imaging mechanisms, radiometric properties, and noise characteristics. Directly decoding such fused features into LULC predictions may not fully resolve the semantic gap between modality-specific representations and the final land cover categories. To further organize heterogeneous features in a task-oriented semantic space, we introduce a USMT after HIAA.

Different from arbitrary modality translation methods that synthesize target-modality images, USMT is not designed for image reconstruction or cloud-free optical image generation. Instead, it performs semantic latent mapping for LULC prediction. In other words, the role of USMT is to transform the fused SAR–optical tokens into a compact semantic representation that is more directly aligned with land cover discrimination. Although USMT adopts a Diffusion Transformer (DiT) -style token-based transformer structure, it does not perform diffusion-based generation or denoising. The term DiT-style here refers to the use of transformer blocks to process latent tokens rather than convolutional feature maps.

Given the fused semantic tokens produced by HIAA, denoted as $\mathbf{T}^f$, USMT first applies a linear embedding layer and positional encoding to preserve token-level spatial organization:

$$\mathbf{Z}^{0} = \mathbf{T}^{f}\mathbf{W}_{e} + \mathbf{E}_{pos} \tag{24}$$

where $\mathbf{W}_e$ denotes a learnable embedding projection, and $\mathbf{E}_{pos}$ denotes positional encoding. The resulting token sequence $\mathbf{Z}^0$ is then processed by a stack of $L$ DiT-style transformer blocks. The embedded token sequence $\mathbf{Z}^0$ is then processed by a stack of $L$ DiT-style transformer blocks, where $L$ is fixed to 4 in all experiments. For the $l$-th block, the transformation can be written as:

$$\mathbf{Z}^{l} = \mathrm{USMT}_{l}(\mathbf{Z}^{l-1}), \qquad l = 1,2,\cdots,L \tag{25}$$

Each USMT block follows a transformer-style token mixing structure consisting of normalization, multi-head self-attention, feed-forward transformation, and residual connections (Vaswani et al., 2017). Following the latent-token modeling design of DiT (Peebles and Xie, 2023), USMT uses transformer blocks to model semantic latent tokens, but its objective is dense LULC semantic mapping rather than image generation. This token-based mapping enables the model to capture long-range semantic dependencies among fused SAR–optical tokens and organize them in a unified LULC-oriented latent space.

The output of the final transformer block is regarded as the predicted semantic latent representation:

$$\hat{\mathbf{Z}} = \mathbf{Z}^{L} \tag{26}$$

where $\hat{\mathbf{Z}} \in \mathbb{R}^{M \times d_z}$ denotes the predicted semantic latent representation. This semantic latent representation serves two purposes. First, it provides a structured intermediate representation for the segmentation decoder. Second, during training, it is constrained by the label-derived semantic anchor described in Section 4.5, which encourages the fused SAR–optical features to be aligned with the LULC semantic space rather than being optimized only through pixel-wise classification.

The predicted semantic latent representation $\hat{\mathbf{Z}}$ is subsequently decoded into the class-probability map $\mathbf{P}$ and the final LULC map $\hat{\mathbf{Y}}$ following the prediction process defined in Eq. (6) and Eq. (7). By introducing USMT, CloudLULC-Net separates heterogeneous feature aggregation from semantic latent mapping. This design allows HIAA to focus on SAR–optical spatial–channel interaction, while USMT further transforms the aggregated representation into a unified LULC semantic space for robust cloud-contaminated LULC mapping.

### 4.5. Semantic anchor-guided optimization

Although the class-probability map provides direct pixel-level supervision for LULC prediction, relying only on segmentation loss may be insufficient to constrain the intermediate semantic representation learned by USMT. In cloud-contaminated SAR–optical mapping, heterogeneous features may contain incomplete optical responses, SAR-specific structural patterns, and modality-dependent noise. If the semantic latent representation is supervised only through the final classification objective, the network may learn latent tokens that are discriminative for training samples but insufficiently structured for robust generalization across regions and cloud conditions. To address this issue, we introduce a semantic anchor-guided optimization strategy to explicitly constrain the predicted semantic latent representation.

Given the ground-truth LULC label map $\mathbf{Y}$, a lightweight semantic anchor encoder $E_y(\mathbf{Y})$ is used to transform the discrete label map into a semantic anchor embedding:

$$\mathbf{Z}_y = E_y(\mathbf{Y}) \tag{27}$$

where $\mathbf{Z}_y \in \mathbb{R}^{M \times d_z}$ denotes the label-derived semantic anchor, $M$ denotes the number of semantic anchor tokens, and $d_z$ denotes the latent embedding dimension. The anchor encoder converts pixel-level categorical supervision into a compact latent representation, providing a structured semantic target for the predicted semantic latent $\hat{\mathbf{Z}}$ generated by USMT. This design is related to recent latent anchor regression paradigms for remote sensing modality mapping, where anchor constraints are used to stabilize latent-space learning (Chen et al., 2026). However, unlike modality translation methods that use latent anchors for target image synthesis, the semantic anchor in CloudLULC-Net is derived from LULC labels and is used to regularize semantic representation learning for segmentation.

The predicted semantic latent $\hat{\mathbf{Z}}$ is aligned with the label-derived semantic anchor $\mathbf{Z}_y$ through an anchor regression loss:

$$\mathcal{L}_{anchor} = \frac{1}{M}\sum_{m=1}^{M}\left\|\hat{\mathbf{Z}}_m - \mathbf{Z}_{y,m}\right\|_2^2 \tag{28}$$

where $\hat{\mathbf{Z}}_m$ and $\mathbf{Z}_{y,m}$ denote the $m$-th predicted semantic token and the corresponding anchor token, respectively. This loss encourages the semantic latent predicted from cloud-contaminated SAR–optical observations to be aligned with the semantic structure encoded from the ground-truth LULC map. In this way, the network learns a more organized LULC-oriented latent space rather than relying only on local pixel-wise classification supervision.

For dense LULC prediction, the class-probability map $\mathbf{P}$ generated by the decoder is supervised using a segmentation loss. Pixel-wise cross-entropy is widely used for semantic segmentation because it directly optimizes per-pixel class prediction, while Dice loss is commonly introduced to improve region-level overlap and alleviate the influence of class imbalance (Long et al., 2015; Milletari et al., 2016; Yeung et al., 2022). Therefore, the segmentation loss in this study combines cross-entropy loss and Dice loss:

$$\mathcal{L}_{seg} = \mathcal{L}_{CE} + \mathcal{L}_{Dice} \tag{29}$$

where $\mathcal{L}_{CE}$ measures pixel-wise classification accuracy, and $\mathcal{L}_{Dice}$ helps alleviate class imbalance by optimizing region-level overlap. Dice loss is particularly useful for LULC mapping because different land cover categories often exhibit imbalanced spatial distributions across regions.

The final training objective is defined as:

$$\mathcal{L}_{total} = \mathcal{L}_{seg} + \lambda_a \mathcal{L}_{anchor} \tag{30}$$

where $\lambda_a$ controls the contribution of the semantic anchor regression loss and is set to 0.1 in all experiments. During optimization, the segmentation loss directly constrains the final class-probability map, while the anchor regression loss regularizes the intermediate semantic latent representation. This joint objective enables CloudLULC-Net to learn both pixel-level discriminative predictions and structured LULC-oriented semantic representations.

## 5. Experiments and analysis

This section evaluates the effectiveness of the proposed CloudLULC-Net on the CloudLULC-Set benchmark. The experiments are organized to examine both the accuracy and practical applicability of the proposed framework for cloud-contaminated SAR–optical LULC mapping. First, representative methods are compared under a unified experimental protocol to assess the overall classification performance and computational efficiency of CloudLULC-Net. Then, comparative validation with existing global LULC products is conducted to further examine the potential of the proposed method for near-real-time land cover mapping under cloud-contaminated observation conditions.

### 5.1 Experimental design and evaluation metrics

#### 5.1.1 Implementation details

All experiments were conducted on CloudLULC-Set. The dataset was divided into training, validation, and test sets with a ratio of 8:1:1, and the same split was used for all compared methods to ensure fair evaluation. Unless otherwise specified, all compared methods were retrained under the same dataset split and training protocol.

CloudLULC-Net was implemented using the PyTorch deep learning framework and trained on a workstation equipped with a single NVIDIA RTX A5000 GPU. The model was optimized using the AdamW optimizer with $\beta_1 = 0.9$, $\beta_2 = 0.99$, and a weight decay of $1 \times 10^{-4}$. The initial learning rate was set to $1 \times 10^{-4}$ and gradually decayed to $1 \times 10^{-6}$ using a cosine annealing strategy. The mini-batch size was set to 12, and the model was trained for 60 epochs. Standard data augmentation operations, including random flipping and scale variation, were applied to improve model generalization.

5.1.2 Evaluation metrics

To evaluate LULC mapping performance, Overall Accuracy (OA), F1-score (F1), and mean Intersection over Union (mIoU) were adopted as quantitative metrics. OA reflects the overall proportion of correctly classified pixels, F1-score measures the balance between precision and recall, and mIoU evaluates the average overlap between predicted and reference regions across all land cover categories. Considering the class imbalance in LULC mapping, mIoU is used as the primary accuracy indicator.

In addition to classification accuracy, computational efficiency was evaluated using FLOPs, parameter number (Param.), and frames per second (FPS), which jointly characterize computational cost, model complexity, and inference efficiency. For a fair comparison, these efficiency metrics were measured using $512 \times 512$ input images on the same GPU platform.

**5.2 Comparative Results**

5.2.1 Comparison with representative methods

To comprehensively evaluate the effectiveness of CloudLULC-Net, we compared it with representative methods under the same CloudLULC-Set benchmark protocol. Following the heterogeneous SAR–optical paradigms illustrated in Fig. 1(c) and Fig. 1(d), the compared methods were divided into two groups: heterogeneous reconstruction-first mapping methods and heterogeneous end-to-end mapping methods. The former first reconstructs cloud-contaminated optical images with the assistance of SAR observations and then performs LULC classification on the reconstructed optical images. The latter directly predicts LULC maps from cloud-contaminated optical and SAR observations within a unified model, without explicitly generating cloud-free optical imagery.

For the heterogeneous reconstruction-first mapping strategy, three SAR-assisted optical reconstruction methods were selected, including HDRSA-Net (Pan et al., 2024), SCT-CR (Ma et al., 2024a), and SAR-DeCR (Wang et al., 2025). HDRSA-Net employs hybrid dynamic residual self-attention to exploit SAR-derived structural priors for optical cloud and shadow removal. SCT-CR introduces a synergistic convolution-transformer structure to enhance SAR–optical feature collaboration for cloud removal. SAR-DeCR is a diffusion-based SAR-fused thick cloud removal method that combines diffusion modeling and Transformer-based representation learning to recover cloud-obscured optical information. The cloud-restored optical images generated by these methods were classified using the same SegFormer-B2 segmentation backbone (Xie et al., 2021), so that the comparison mainly reflects the influence of the reconstruction-first strategy rather than differences in the classifier.

For the heterogeneous end-to-end mapping strategy, eleven representative SAR–optical fusion or multimodal semantic segmentation methods were selected, including SO_fusion (Chen and Bruzzone, 2021), CMX (Zhang et al., 2023), ASANet (Zhang et al., 2024), CloudSeg (Xu et al., 2024), CMFFNet (Guo et al., 2024), FTransUNet (Ma et al., 2024b), SoftFormer (Liu et al., 2024), SpecSAR-Former (Yu et al., 2024), MFFNet (Li et al., 2025a), SegCR (Wu et al., 2025), and STSNet (Yu et al., 2025). SO_fusion exploits SAR–optical synergy through self-supervised representation learning. CMX provides a generic RGB-X semantic segmentation framework with cross-modal feature rectification and fusion. ASANet introduces asymmetric semantic alignment for RGB–SAR land cover classification. CloudSeg is designed for robust land cover mapping under cloudy conditions through multimodal learning. CMFFNet exploits cross-modal multi-scale feature fusion for optical–SAR land surface mapping. FTransUNet, SoftFormer, and SpecSAR-Former use Transformer-based structures to enhance long-range dependency modeling in multimodal remote sensing segmentation. MFFNet incorporates wavelet-based multimodal frequency fusion to enhance complementary feature aggregation for remote sensing semantic segmentation. SegCR introduces a multimodal and multitask complementary fusion framework to reduce the negative influence of cloud interference in optical–SAR semantic segmentation. STSNet incorporates spatial, temporal, and spectral cues for high-resolution land-cover segmentation.

As reported in Table 3, CloudLULC-Net achieves the best overall classification performance among all compared methods, with an OA of 86.60%, an F1-score of 83.29%, and an mIoU of 73.51%. Compared with the strongest baseline SegCR, CloudLULC-Net improves OA, F1-score, and mIoU by 0.94%, 1.25%, and 0.78%, respectively. Compared with ASANet, which also performs competitively by explicitly modeling RGB–SAR semantic alignment, the proposed method still obtains gains of 1.31%, 1.83%, and 1.43% in OA, F1-score, and mIoU, respectively. These improvements indicate that the proposed ORM, HIAA, USMT, and semantic anchor-guided optimization can more effectively handle the semantic uncertainty introduced by cloud-contaminated optical observations. In addition, the reconstruction-first methods achieve reasonable performance after SAR-assisted optical restoration, but they remain inferior to the strongest end-to-end methods, suggesting that visual reconstruction does not necessarily preserve all class-discriminative information required for LULC classification. In terms of computational efficiency, CloudLULC-Net reaches the highest inference speed of 42.27 FPS, while maintaining moderate FLOPs and parameter numbers. Compared with SegCR, CloudLULC-Net reduces FLOPs from 37.62 G to 20.51 G and improves inference speed from 20.83 FPS to 42.27 FPS. These results demonstrate that CloudLULC-Net provides a favorable balance between classification accuracy and computational efficiency, making it suitable for large-scale cloud-contaminated SAR–optical LULC mapping.

**Table 3**
Quantitative comparison of CloudLULC-Net and representative baseline models in terms of classification performance (OA, F1 score, mIoU) and computational efficiency (FLOPs, Param., FPS). To ensure fair comparison, all models are evaluated under a unified setting using 512×512 input images and an NVIDIA RTX A5000 GPU. The highest performance scores are highlighted in bold, and the second-best results are indicated with underlining.

| Method | Paradigm | OA (%) | F1 (%) | mIoU (%) | FLOPs (G) | Param. (M) | Speed (FPS) |
|---|---|---|---|---|---|---|---|
| HDRSA-Net (Pan et al., 2024) | reconstruction-first | 83.23 | 78.60 | 69.18 | 94.72 | 66.84 | 8.74 |
| SCT-CR (Ma et al., 2024a) | reconstruction-first | 82.26 | 77.49 | 68.16 | 108.36 | 72.51 | 7.92 |
| SAR-DeCR (Wang et al., 2025) | reconstruction-first | 84.08 | 79.78 | 70.59 | 188.64 | 118.32 | 3.18 |
| SO_fusion (Chen and Bruzzone, 2021) | end-to-end | 72.43 | 64.41 | 58.64 | 58.95 | 207.72 | 12.57 |
| CMX (Zhang et al., 2023) | end-to-end | 73.38 | 68.84 | 59.97 | 14.35 | 66.87 | 26.69 |
| ASANet (Zhang et al., 2024) | end-to-end | 85.29 | 81.46 | 72.08 | 23.20 | 83.41 | 35.88 |
| CloudSeg (Xu et al., 2024) | end-to-end | 80.70 | 75.37 | 68.50 | 46.17 | 283.93 | 10.00 |
| CMFFNet (Guo et al., 2024) | end-to-end | 79.90 | 74.53 | 67.16 | 31.84 | 52.37 | 24.91 |
| FTransUNet (Ma et al., 2024b) | end-to-end | 78.94 | 73.26 | 66.87 | 40.19 | 166.65 | 18.61 |
| SoftFormer (Liu et al., 2024) | end-to-end | 83.16 | 78.93 | 71.00 | 24.88 | 34.76 | 32.54 |
| SpecSAR-Former (Yu et al., 2024) | end-to-end | 82.44 | 78.27 | 69.68 | 18.96 | 21.43 | 39.18 |
| MFFNet (Li et al., 2025a) | end-to-end | 83.90 | 79.51 | 71.24 | 27.74 | 41.06 | 29.67 |
| SegCR (Wu et al., 2025) | end-to-end | <u>85.66</u> | <u>82.04</u> | <u>72.73</u> | 37.62 | 88.54 | 20.83 |
| STSNet (Yu et al., 2025) | end-to-end | 81.75 | 77.74 | 70.35 | 29.03 | 16.29 | 20.15 |
| CloudLULC-Net (ours) | end-to-end | **86.60** | **83.29** | **73.51** | 20.51 | 25.90 | 42.27 |

To further analyze the category-level performance of different methods, Fig. 6 reports the class-wise IoU and F1-score for six LULC categories. Overall, CloudLULC-Net achieves the most balanced performance across different land cover types. For categories with relatively clear spatial or scattering characteristics, such as water and forest, most methods achieve comparatively high scores, while CloudLULC-Net further improves the IoU of water and forest to 92.61% and 84.21%, respectively. This indicates that the proposed framework can effectively exploit the complementary information between optical spectral responses and SAR structural cues. For more easily confused categories, such as cropland, built-up land, and barren land, CloudLULC-Net also shows consistent advantages. Specifically, it achieves IoU values of 72.21%, 61.84%, and 66.56% for cropland, built-up land, and barren land, respectively, outperforming the strongest baseline SegCR by 0.76%, 0.50%, and 0.66%. Similar trends can also be observed in the F1-score comparison, where CloudLULC-Net obtains the highest F1-scores for forest, cropland, built-up land, barren land, and water. Although CloudSeg shows slightly better performance on grassland, its performance on other categories is less stable, resulting in lower overall mIoU and mean F1-score. These results suggest that CloudLULC-Net does not merely improve the overall accuracy but also enhances the discrimination of heterogeneous and easily confused land cover categories under cloud-contaminated observation conditions.

The qualitative results in Fig. 7 further confirm the advantages of CloudLULC-Net under different cloud-contaminated scenarios. The selected examples contain diverse land cover compositions, including water, forest, grassland, barren land, built-up land, and cropland, as well as different levels of cloud and shadow interference. Compared with SO_fusion and FTransUNet, the proposed method produces more spatially continuous and semantically consistent LULC maps. These conventional fusion methods tend to generate fragmented predictions in cloud-contaminated regions, especially for cropland, built-up land, and barren land, where spectral responses are easily distorted by clouds and shadows. CloudSeg improves the robustness to cloud interference to some extent, but its predictions still contain local discontinuities and category confusion in heterogeneous regions. SAR-DeCR, as a reconstruction-first method, can recover certain optical structures before classification; however, the reconstructed optical images may still introduce semantic distortion or lose class-discriminative details, leading to inaccurate boundaries and local misclassification. ASANet and SegCR generate more competitive results by exploiting SAR–optical complementary information, but they still show confusion among cropland, built-up land, and barren land in several complex scenes. In contrast, CloudLULC-Net better preserves the spatial structure of water bodies, maintains more complete forest and cropland regions, and produces clearer boundaries between built-up land and surrounding land cover types. These visual comparisons indicate that the proposed framework can effectively combine reliability-aware optical modulation, heterogeneous high-order interaction, and semantic latent mapping to generate more reliable LULC maps from cloud-contaminated optical and SAR observations.

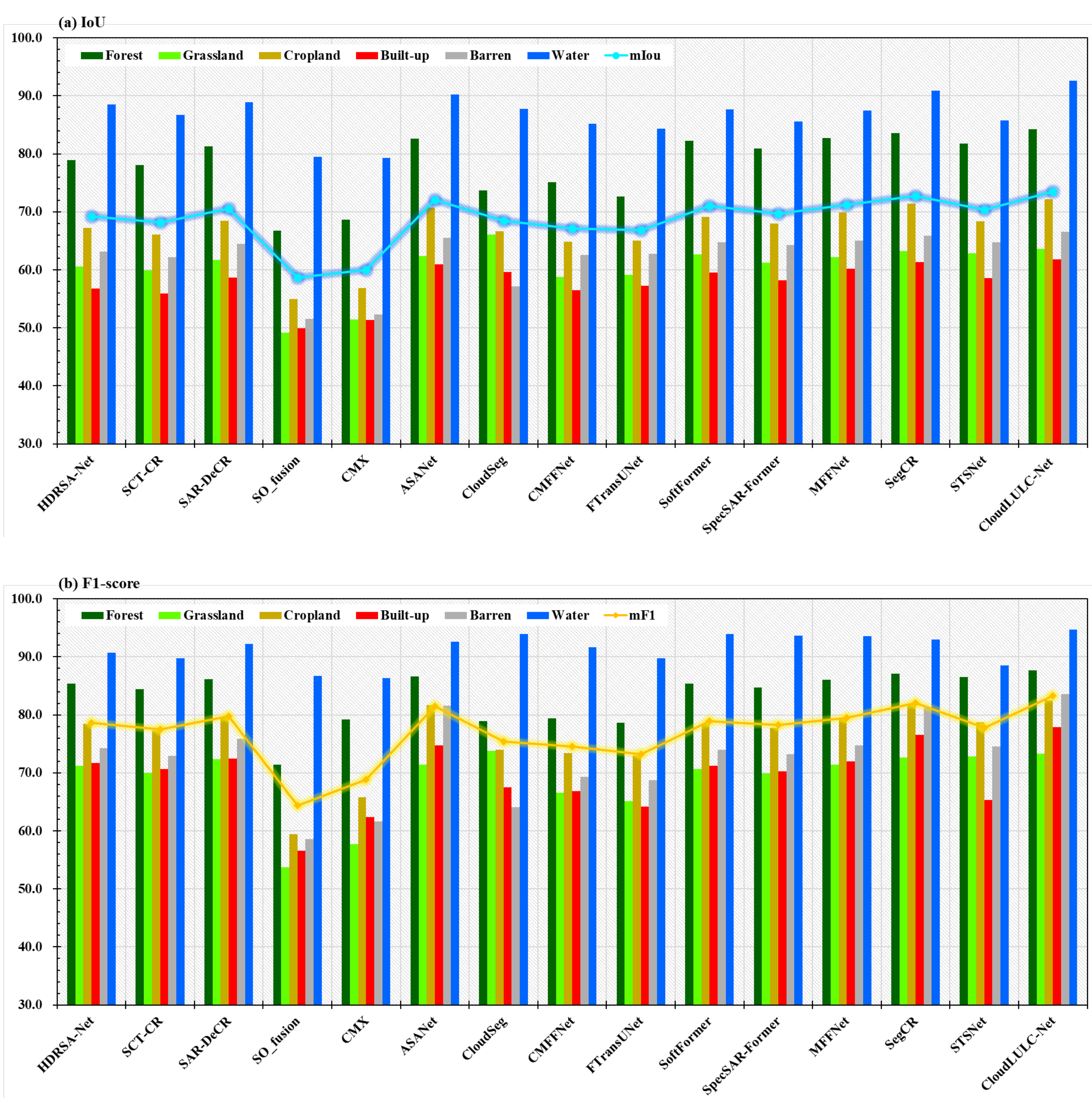

**Fig. 6.** Class-wise comparison of CloudLULC-Net and representative methods on CloudLULC-Set. (a) IoU of different LULC categories and mIoU. (b) F1-score of different LULC categories and mean F1-score.

### 5.2.2 Comparative validation with existing global LULC products

To further evaluate the practical value of CloudLULC-Net for target-date LULC mapping under cloud-contaminated conditions, we compared the generated maps with representative global LULC products in two geographically distinct regions: Maryland, United States, and Tamil Nadu, India, whose geographical locations are shown in Fig. 2. The cloud-contaminated Sentinel-2 observations acquired on August 28, 2021 for Maryland and February 7, 2023 for Tamil Nadu were used as target scenes, each covering an area of approximately $109.8 \times 109.8\ \text{km}^2$. This experiment was designed as an external product-level validation rather than a strictly equivalent supervised model comparison, because existing global LULC products differ from CloudLULC-Net in temporal resolution, production strategy, data sources, and class definitions.

For the Maryland region, three widely used global LULC products were included for comparison, namely Esri Land Cover (Karra et al., 2021), ESA WorldCover (Zanaga et al., 2022), and Google Dynamic World (Brown et al., 2022). For the Tamil Nadu region, Esri Land Cover, GLC_FCS10 (Zhang et al., 2025b), and Google Dynamic World were used. Esri Land Cover, ESA WorldCover, and GLC_FCS10 are annual or product-level global LULC datasets, whereas Google Dynamic World provides near-real-time land cover probabilities based on Sentinel-2 observations. To ensure class-level comparability, all product labels were harmonized into the six-class taxonomy used in this study, including water, forest, grassland, barren land, built-up land, and cropland. The products were also resampled and aligned to the same spatial grid before quantitative evaluation. Reference labels were manually interpreted by experts with the assistance of temporally adjacent cloud-free or near-cloud-free Sentinel-2 observations.

Table 4 presents the quantitative comparison in the Maryland region. CloudLULC-Net achieves the best overall performance, with an OA of 93.16%, an F1-score of 76.59%, and an mIoU of 73.47%. Compared with the strongest global product in this region, ESA WorldCover, CloudLULC-Net improves OA, F1-score, and mIoU by 1.52%, 3.10%, and 4.26%, respectively. In terms of class-wise IoU, the proposed method also achieves the highest accuracy for all six LULC categories. The improvements are particularly evident for barren land, built-up land, cropland, and grassland, indicating that CloudLULC-Net can better represent local heterogeneous land cover patterns under cloud-contaminated target-date observations.

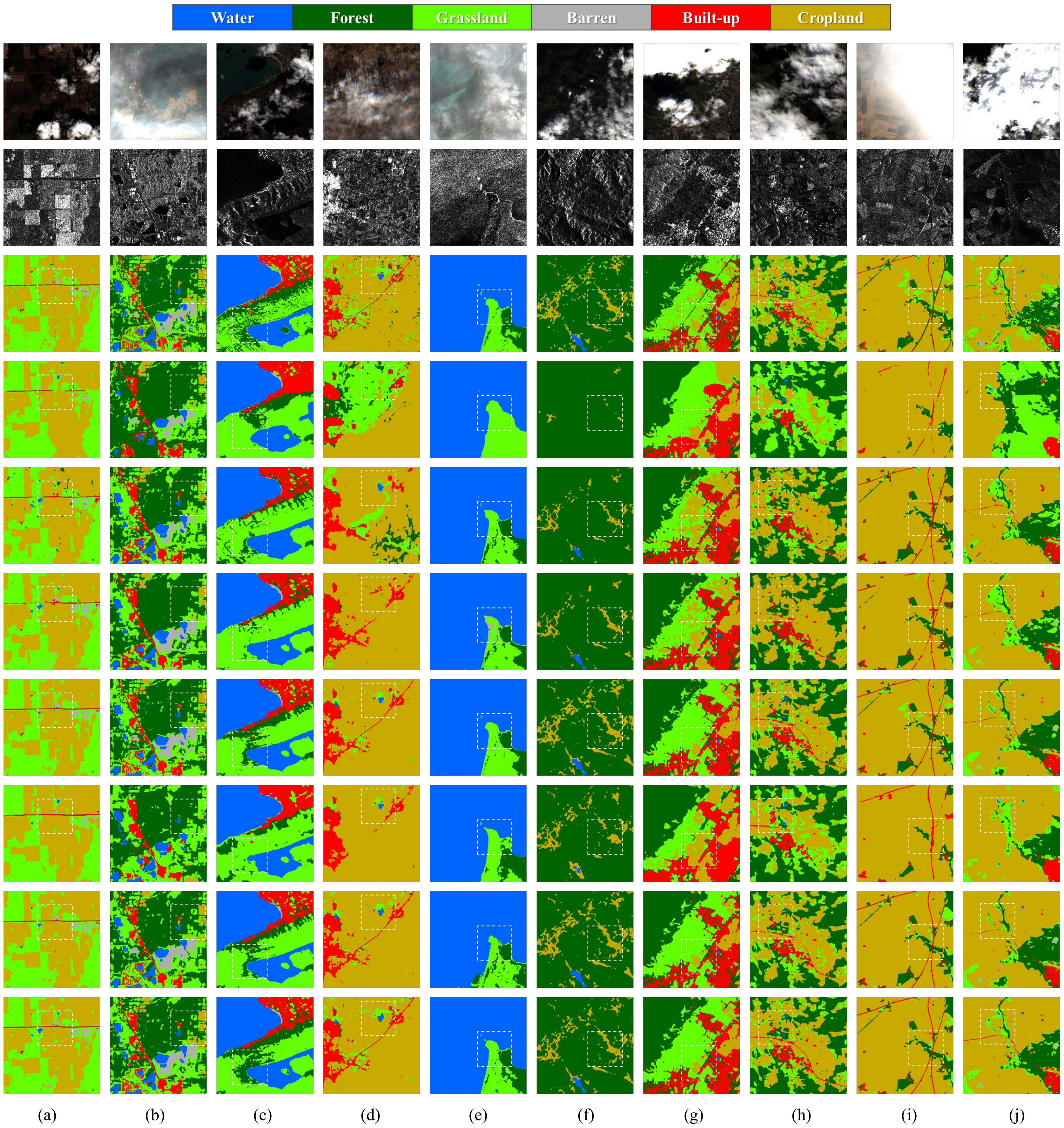

**Fig. 7.** Visual comparison of LULC mapping results generated by representative methods on CloudLULC-Set. The first three rows show the cloud-contaminated optical images, corresponding SAR images, and reference LULC labels. Starting from the fourth row, the results generated by SO_fusion, FTransUNet, CloudSeg, SAR-DeCR, ASANet, SegCR, and the proposed CloudLULC-Net are presented sequentially.

**Table 4**

Quantitative comparison of LULC mapping performance between CloudLULC-Net and representative global LULC products in the Maryland region, United States. Bold values indicate the best performance, and underlined values denote the second-best.

| Global LULC product | OA (%) | F1 (%) | mIoU (%) | IoU (%) | | | | | |
|---|---|---|---|---|---|---|---|---|---|
| | | | | Water | Forest | Barren | Built-up | Cropland | Grassland |
| Esri Land Cover (Karra et al., 2021) | 84.55 | 62.07 | 51.42 | 97.11 | 79.72 | 11.74 | 19.77 | 63.98 | 36.18 |
| ESA WorldCover (Zanaga et al., 2022) | <u>91.64</u> | <u>73.49</u> | <u>69.21</u> | <u>97.15</u> | <u>88.62</u> | <u>51.26</u> | <u>52.97</u> | <u>74.91</u> | <u>50.37</u> |
| Google Dynamic World (Brown et al., 2022) | 74.93 | 46.00 | 36.73 | 91.18 | 68.06 | 0.55 | 21.03 | 29.72 | 9.86 |
| CloudLULC-Net (ours) | **93.16** | **76.59** | **73.47** | **98.17** | **90.41** | **58.89** | **56.11** | **79.13** | **58.10** |

**Table 5**

Quantitative comparison of LULC mapping performance between CloudLULC-Net and representative global LULC products in the Tamil Nadu region, India. Bold values indicate the best performance, and underlined values denote the second-best.

| Global LULC product | OA (%) | F1 (%) | mIoU (%) | IoU (%) | | | | | |
|---|---|---|---|---|---|---|---|---|---|
| | | | | Water | Forest | Barren | Built-up | Cropland | Grassland |
| Esri Land Cover (Karra et al., 2021) | 81.72 | 58.74 | 47.00 | 61.01 | 76.73 | 5.80 | 25.79 | 80.51 | 32.15 |
| GLC_FCS10 (Zhang et al., 2025b) | 82.39 | 54.68 | 44.32 | 45.90 | 85.41 | 0.00 | 56.48 | 66.38 | 11.77 |
| Google Dynamic World (Brown et al., 2022) | 73.91 | 46.53 | 34.53 | 40.55 | 72.83 | 6.03 | 30.05 | 51.55 | 6.17 |
| CloudLULC-Net (ours) | **92.33** | **79.63** | **69.27** | **75.95** | **92.24** | **32.19** | **79.57** | **88.06** | **47.59** |

Table 5 reports the results in the Tamil Nadu region, where the landscape is more complex and the target optical observation is more severely affected by clouds. CloudLULC-Net again achieves the best performance, with an OA of 92.33%, an F1-score of 79.63%, and an mIoU of 69.27%. Compared with the best-performing global products in terms of each overall metric, CloudLULC-Net improves OA by 9.94% over GLC_FCS10, F1-score by 20.89% over Esri Land Cover, and mIoU by 22.27% over Esri Land Cover. The class-wise results also show clear advantages, especially for water, barren land, built-up land, cropland, and grassland. These results suggest that the proposed method has stronger adaptability to target-date LULC mapping in tropical and agriculturally complex regions, where global products may suffer from temporal mismatch, cloud-related missing observations, or insufficient local detail.

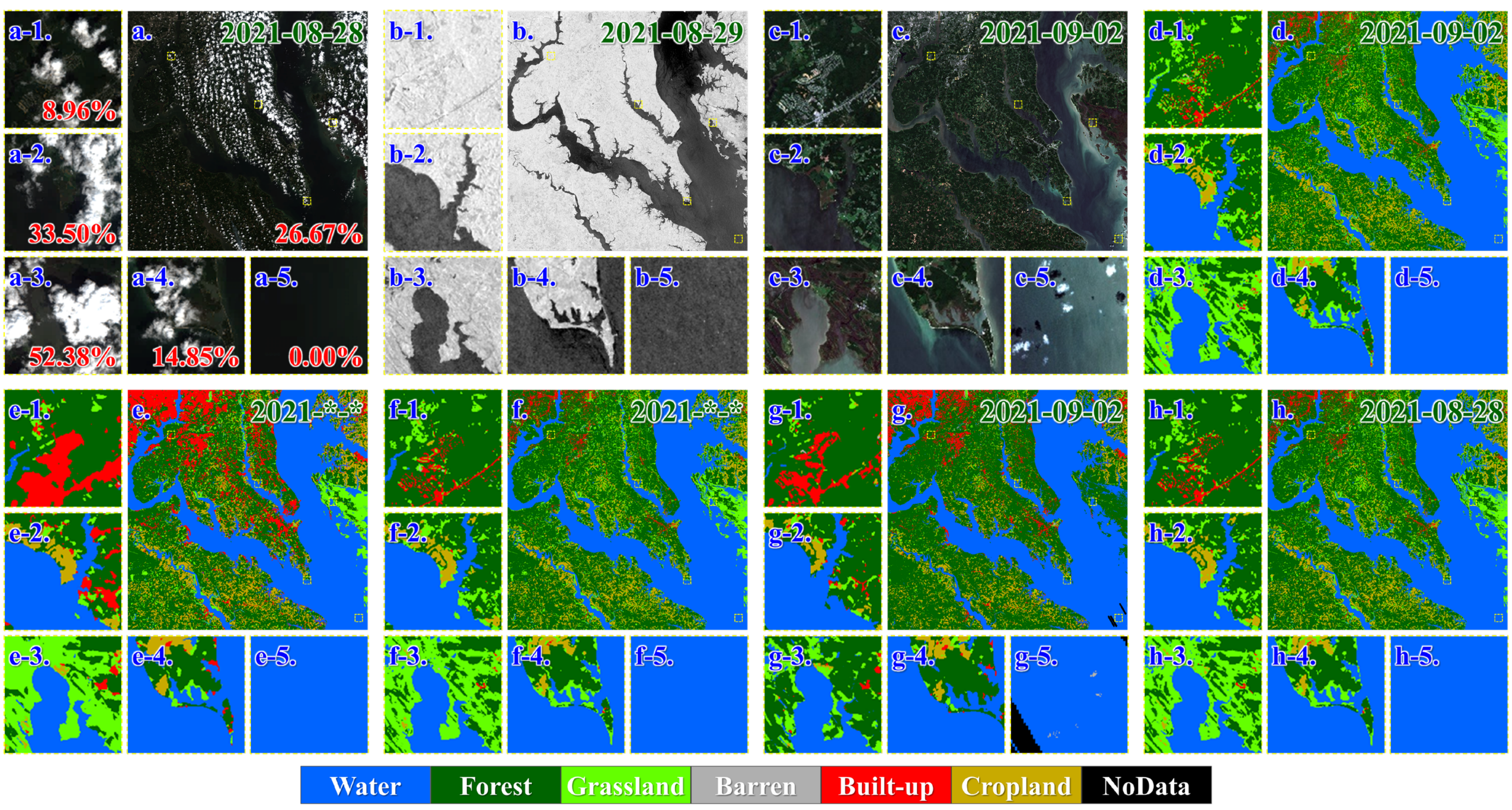

**Fig. 8.** Visual comparison of LULC mapping results in the Maryland region with corresponding local magnifications. The upper panels show (a) cloud-contaminated optical imagery, (b) the corresponding SAR image, (c) the nearest cloud-free optical image used for reference interpretation, and (d) manually annotated reference labels. The lower panels show the LULC maps generated by (e) Esri Land Cover, (f) ESA WorldCover, (g) Google Dynamic World, and (h) the proposed CloudLULC-Net. Red numerals indicate cloud-cover percentages, green numerals denote image acquisition dates, and an asterisk (*) indicates unavailable product timestamp information.

The visual comparisons in Fig. 8 and Fig. 9 further support the quantitative results. In the Maryland region, the global products generally capture large-scale water and forest distributions, but they show noticeable local inconsistencies in fragmented built-up, cropland, and barren areas. ESA WorldCover provides relatively stable spatial patterns, which is consistent with its quantitative performance, but some fine boundaries and local transitions are still not fully represented. Google Dynamic World provides valuable near-real-time land cover information; however, because it relies on valid Sentinel-2 optical observations, its results may become incomplete or fragmented when cloud contamination affects the target period. In contrast, CloudLULC-Net produces maps that are more consistent with the manually interpreted reference labels, especially in the magnified regions where shoreline structures, forest boundaries, and mixed built-up–cropland patterns are better preserved.

Similar observations can be found in the Tamil Nadu region shown in Fig. 9. The target scene contains severe cloud contamination and complex mixtures of cropland, forest, built-up land, and barren land. Esri Land Cover and GLC_FCS10 provide useful large-scale references but tend to smooth local heterogeneous patterns or miss small fragmented patches. Google

Dynamic World shows obvious missing or inconsistent areas in several local regions, reflecting the difficulty of optical-only near-real-time mapping under cloudy conditions. By jointly exploiting cloud-contaminated optical imagery and temporally adjacent SAR observations, CloudLULC-Net generates more complete and spatially coherent LULC maps. It better delineates cropland parcels, forest blocks, water bodies, and built-up areas, and its local magnified results are more consistent with the reference annotations.

Overall, the comparison with existing global LULC products demonstrates the practical advantage of CloudLULC-Net for cloud-contaminated target-date LULC mapping. Global LULC products remain valuable for long-term and large-area land cover monitoring, but their temporal granularity, product-generation strategy, and dependence on valid optical observations may limit their ability to represent fine-scale land surface conditions at specific cloudy observation dates. In contrast, CloudLULC-Net directly uses cloud-contaminated Sentinel-2 imagery and temporally adjacent Sentinel-1 SAR observations, enabling more accurate and timely LULC mapping for fine-grained regional analysis.

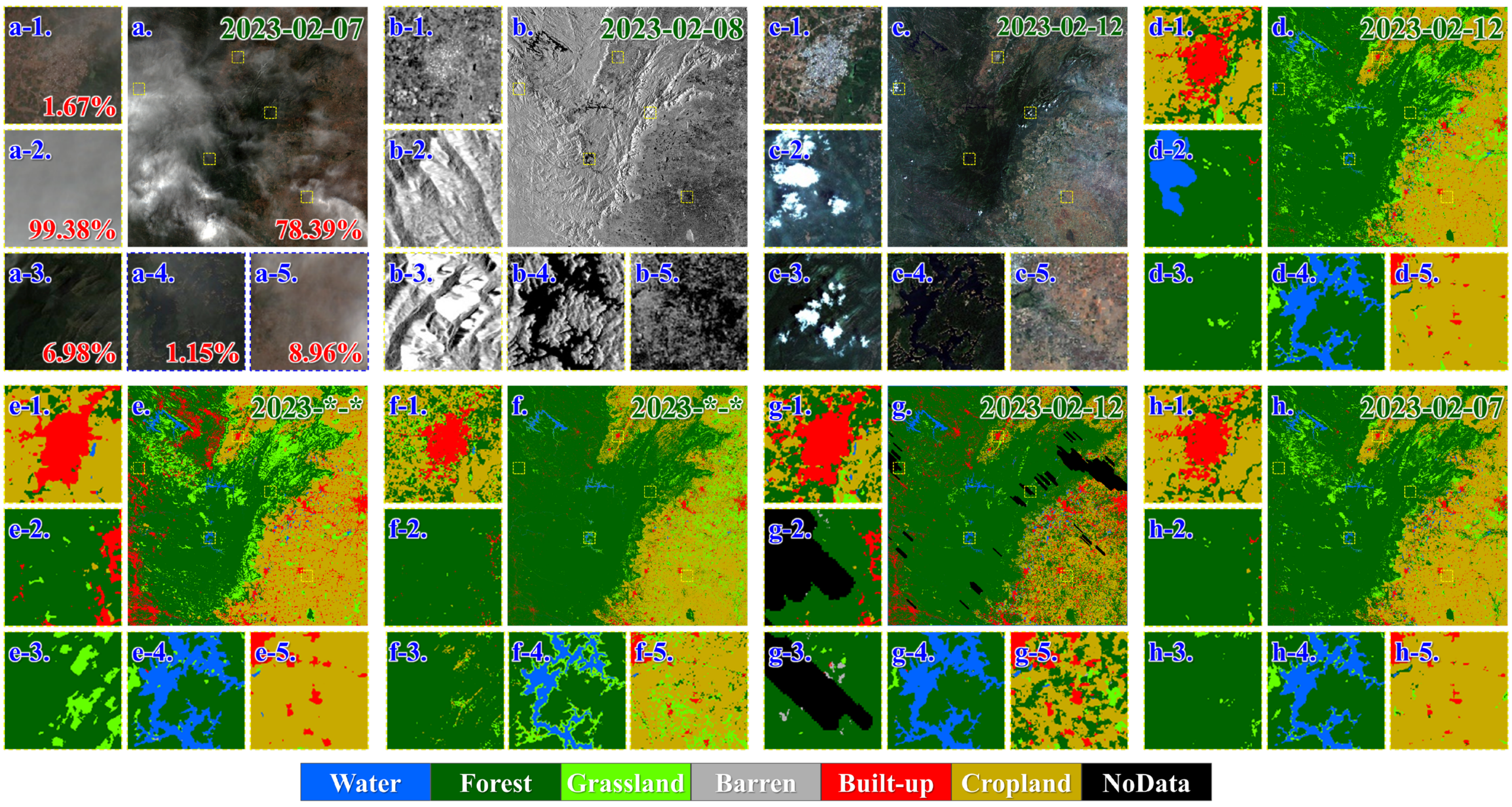

**Fig. 9.** Visual comparison of LULC mapping results in the Tamil Nadu region with corresponding local magnifications. The upper panels show (a) cloud-contaminated optical imagery, (b) the corresponding SAR image, (c) the nearest cloud-free optical image used for reference interpretation, and (d) manually annotated reference labels. The lower panels show the LULC maps generated by (e) Esri Land Cover, (f) GLC_FCS10, (g) Google Dynamic World, and (h) the proposed CloudLULC-Net. Red numerals indicate cloud-cover percentages, green numerals denote image acquisition dates, and an asterisk (*) indicates unavailable product timestamp information.

## 6. Discussion

The experimental results demonstrate that CloudLULC-Net achieves competitive performance for cloud-contaminated SAR–optical LULC mapping. Beyond the overall comparison with representative methods and global LULC products, it is necessary to further analyze why the proposed framework is effective under adverse observation conditions. Therefore, this section discusses the influence of cloud contamination, the contribution of SAR observations, the effectiveness of key modules, and the potential value and limitations of the proposed method for fine-grained spatiotemporal LULC mapping.

### 6.1. Impact of cloud contamination on LULC mapping

Cloud contamination directly affects the reliability of optical-image-based LULC mapping by obscuring land surface reflectance, introducing cloud-shadow interference, and reducing the separability of spectrally similar categories. To analyze the influence of cloud coverage on different methods, we selected the Anhui region in China as a case study, whose geographical location is shown in Fig. 2. The target scene covers an area of approximately $109.8 \times 109.8$ km$^2$. Cloud-covered areas in Sentinel-2 imagery were detected using the Fmask 4.6 tool (Qiu et al., 2019), and the inference samples were grouped into five cloud-coverage intervals: 0–20%, 20–40%, 40–60%, 60–80%, and 80–100%. This setting enables a stratified evaluation of LULC mapping performance under progressively degraded optical observation conditions.

As shown in Fig. 10, the performance of all compared methods generally decreases with increasing cloud coverage, confirming the negative impact of cloud contamination on LULC mapping. Under low-cloud conditions, most methods can still exploit valid optical information and obtain acceptable results. However, as cloud coverage increases, optical spectral and textural cues become incomplete or unreliable, leading to more severe confusion among cropland, grassland, built-up land, and barren land. CloudLULC-Net consistently achieves higher OA, F1-score, and mIoU across all cloud-coverage intervals and exhibits a slower performance decline under medium-to-heavy cloud contamination. Although ASANet and SegCR show stronger robustness than conventional fusion methods, CloudLULC-Net still performs better under most cloud conditions. These results indicate that robust cloud-contaminated LULC mapping requires not only SAR information, but also reliability-aware optical modulation and effective cross-modal semantic interaction.

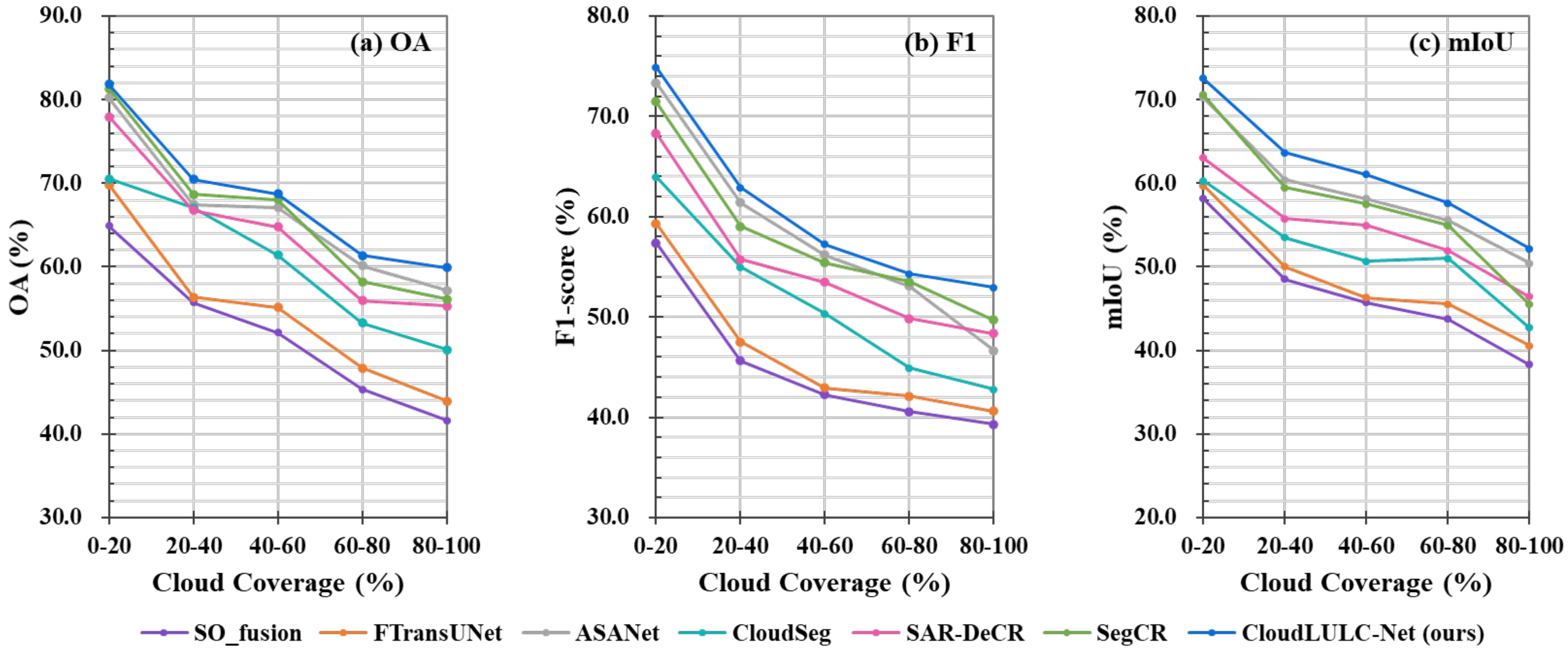

**Fig. 10.** Quantitative comparison of CloudLULC-Net and selected representative methods under different cloud-coverage levels. The selected methods include representative reconstruction-first and end-to-end SAR–optical mapping baselines.

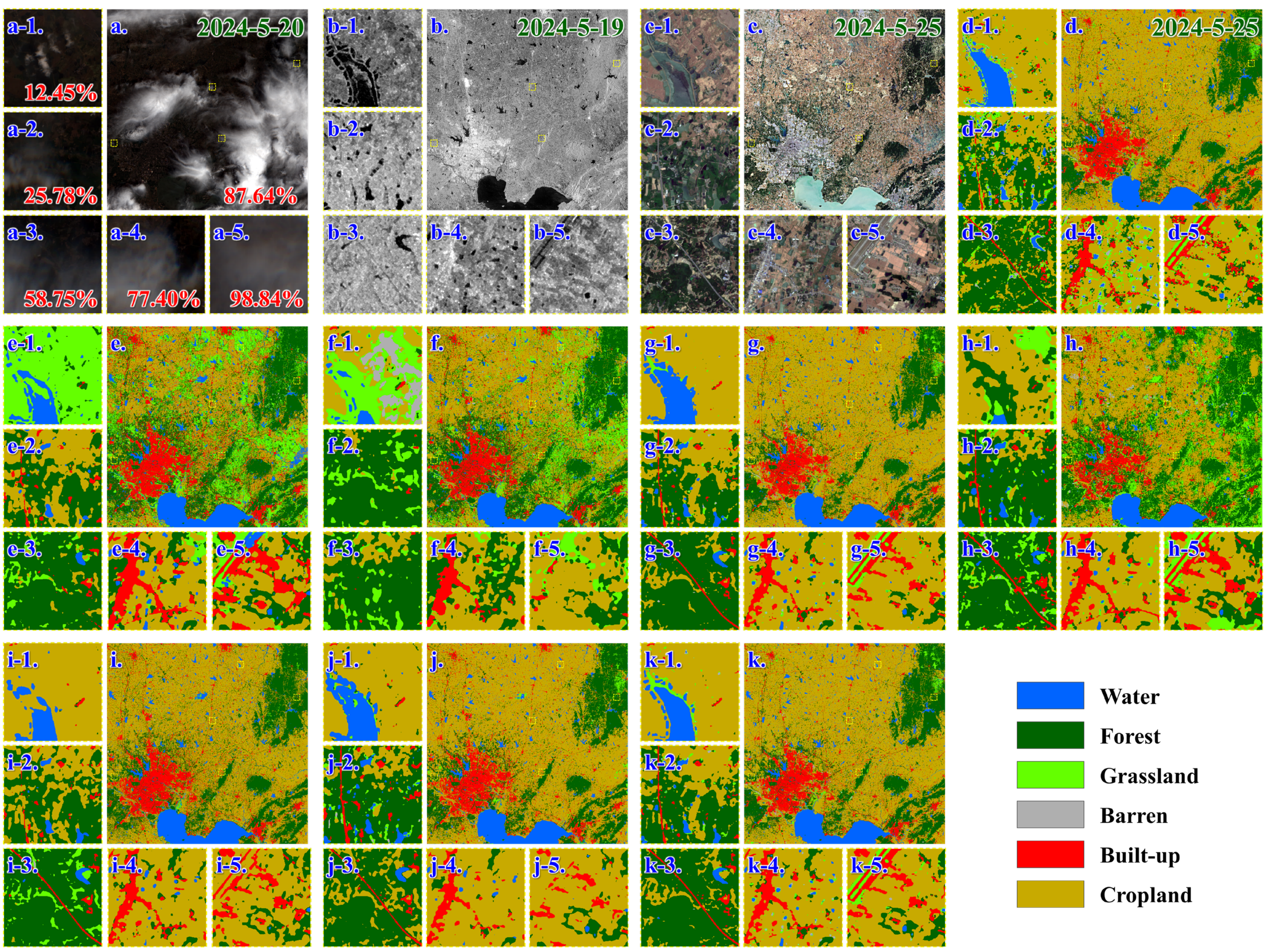

**Fig. 11.** Visual comparison of LULC mapping results generated by CloudLULC-Net and selected representative methods under different cloud-coverage levels, with corresponding local magnifications. Subfigures (a)–(d) show the cloud-contaminated optical images, corresponding SAR images, nearest cloud-free optical images used for reference interpretation, and manually annotated LULC reference labels, respectively. Subfigures (e)–(k) show the LULC mapping results generated by SO_fusion, FTransUNet, CloudSeg, SAR-DeCR, ASANet, SegCR, and CloudLULC-Net, respectively. Red numbers indicate cloud-cover percentages, and green numbers denote image acquisition dates.

The visual comparison in Fig. 11 further supports this observation. As cloud coverage increases, several baselines produce fragmented predictions and local category confusion, especially in cloud-obscured cropland, grassland, built-up, and barren regions. SO_fusion and FTransUNet tend to generate discontinuous land cover patterns, while CloudSeg and SAR-DeCR still suffer from misclassification in heterogeneous areas. ASANet and SegCR provide more competitive results, but their local boundaries remain less stable when optical semantics are severely degraded. In contrast, CloudLULC-Net produces more spatially coherent and semantically consistent maps, with clearer water boundaries, better separation of built-up areas from surrounding cropland and barren land, and fewer fragmented vegetation predictions. This suggests that the proposed framework can effectively compensate for missing or unreliable optical cues by exploiting SAR structural information under heavy cloud contamination.

### 6.2. Contribution of SAR observations to cloud-contaminated LULC mapping

To further analyze the contribution of SAR observations in cloud-contaminated LULC mapping, we evaluated CloudLULC-Net under four different input configurations. The Biobío Region in Chile was selected as the representative test site, whose geographical location is shown in Fig. 2. The target scene covers an area of approximately $109.8 \times 109.8\ \text{km}^2$. Specifically, the four configurations include: (i) cloud-contaminated optical images only, (ii) SAR images only, (iii) temporally adjacent cloud-free optical images only, and (iv) cloud-contaminated optical images combined with SAR images. Among them, the cloud-free optical setting is used as an ideal reference condition to indicate the potential upper-bound performance when valid optical observations are available.

As reported in Table 6, using only cloud-contaminated optical images leads to the lowest classification accuracy, indicating that cloud occlusion severely weakens the discriminative ability of optical spectral and textural features. The SAR-only configuration achieves better robustness in cloud-obscured areas because SAR observations can penetrate clouds and provide structural and backscattering information related to water bodies, built-up areas, and terrain textures. However, SAR imagery alone is still less effective than optical imagery for separating spectrally dependent categories such as cropland, grassland, and forest. As expected, the cloud-free optical configuration achieves high accuracy because it provides complete and reliable surface reflectance information. More importantly, when SAR imagery is combined with cloud-contaminated optical imagery, the classification performance is substantially improved and approaches that obtained under cloud-free optical conditions. This result demonstrates that SAR observations can effectively compensate for missing or unreliable optical cues under cloud contamination.

**Table 6**

Quantitative evaluation of CloudLULC-Net under four different input configurations in the Biobío region, Chile. The symbol "√" indicates that the corresponding data type is used as input, while "–" denotes its absence. Bold values indicate the best performance for each metric, and underlined values denote the second-best.

| Input | | | OA (%) | F1 (%) | mIoU (%) | IoU (%) | | | | | |
|---|---|---|---|---|---|---|---|---|---|---|---|
| Cloudy | SAR | Clear | | | | Water | Forest | Barren | Built-up | Cropland | Grassland |
| √ | - | - | 76.28 | 41.66 | 32.17 | 75.65 | 68.62 | 3.81 | 24.3 | 3.44 | 17.19 |
| - | √ | - | 85.36 | 53.23 | 43.77 | 98.81 | 80.50 | 12.90 | 43.74 | 12.28 | 14.41 |
| - | - | √ | **91.14** | **77.05** | **65.22** | **99.51** | **88.31** | **51.79** | **60.88** | **42.67** | **48.14** |
| √ | √ | - | <u>90.91</u> | <u>76.21</u> | <u>64.29</u> | <u>99.48</u> | <u>88.01</u> | <u>49.75</u> | <u>60.38</u> | <u>40.37</u> | <u>47.77</u> |

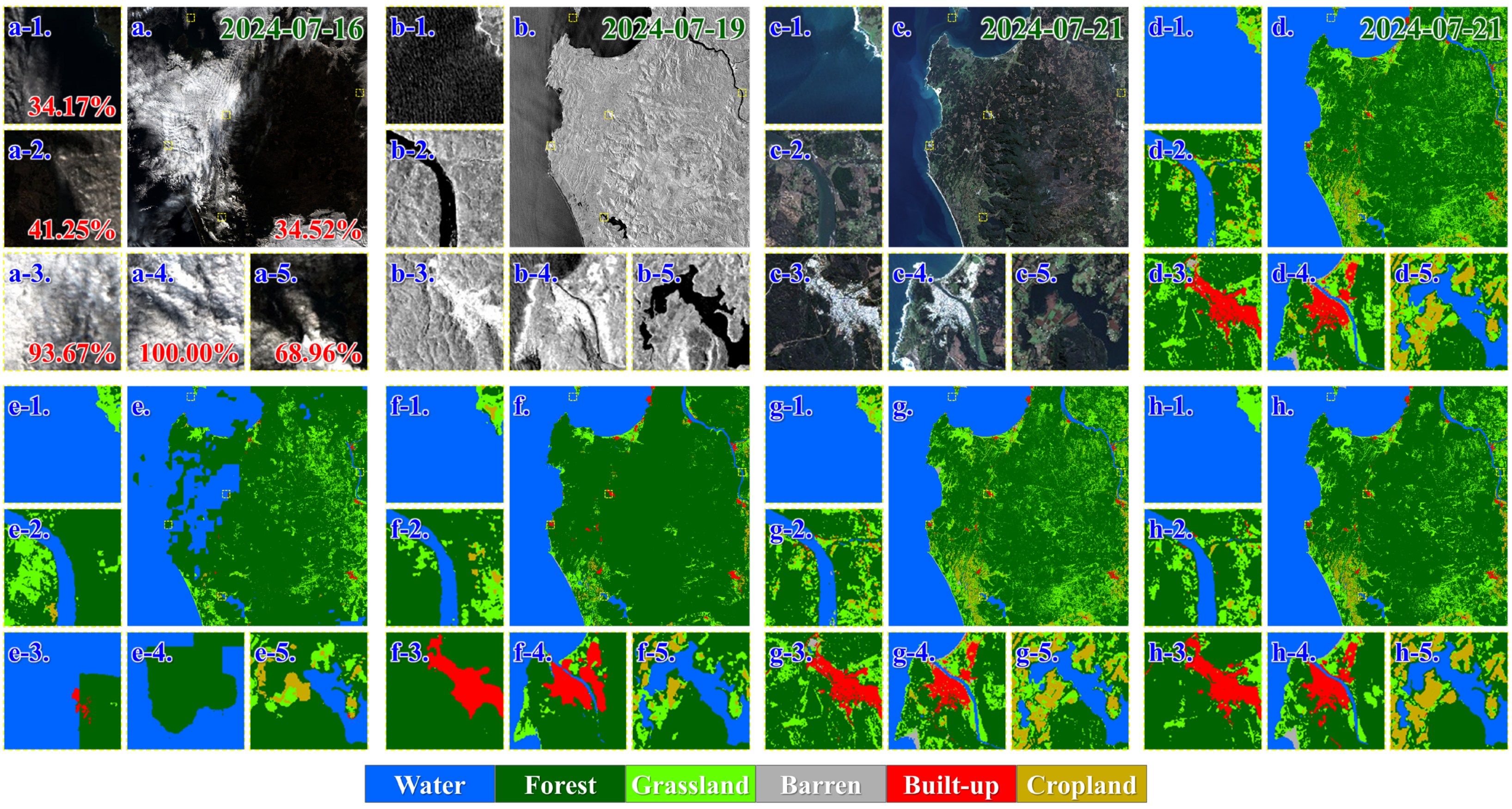

**Fig. 12.** Visual comparison of LULC mapping results generated by CloudLULC-Net under different input configurations in the Biobío region. The upper panels show (a) cloud-contaminated optical imagery, (b) SAR imagery, (c) temporally closest cloud-free optical imagery used as an ideal reference condition, and (d) manually annotated reference labels. The lower panels show the LULC mapping results obtained using (e) cloud-contaminated optical imagery only, (f) SAR imagery only, (g) cloud-free optical imagery only, and (h) cloud-contaminated optical imagery combined with SAR imagery. Red numerals indicate cloud-cover percentages, and green annotations denote image acquisition dates.

The visual comparison in Fig. 12 provides further evidence for this conclusion. When only cloud-contaminated optical images are used, the predicted LULC maps contain obvious omission and misclassification errors in cloud-covered regions, particularly in local areas with dense cloud occlusion. The SAR-only result better preserves some structural patterns, such as water boundaries and built-up textures, but it still shows category confusion in vegetation and agricultural regions. The cloud-free optical result provides more complete semantic information, but such ideal observations are often unavailable at the target date in cloudy regions. In contrast, the combination of cloud-contaminated optical and SAR observations produces more spatially coherent and semantically reliable LULC maps. Water bodies are more clearly delineated, forest and cropland regions are better preserved, and built-up areas are more accurately separated from surrounding land cover types.

These results indicate that the benefit of SAR imagery lies not only in its cloud-penetrating capability, but also in its complementary representation of land surface structure. Under cloud-contaminated conditions, optical imagery still provides valuable spectral information in clear or weakly contaminated regions, while SAR imagery supplies reliable structural cues in cloud-obscured areas. The proposed CloudLULC-Net can integrate these complementary properties and reduce semantic ambiguity caused by cloud contamination, thereby narrowing the performance gap between cloudy observation conditions and ideal cloud-free optical mapping.

**6.3 Ablation study**

6.3.1 Effectiveness of key components

To evaluate the contribution of each key component in CloudLULC-Net, we conducted ablation experiments by progressively introducing the proposed modules into a basic SAR–optical segmentation framework. The baseline model adopts modality-specific encoders and a segmentation decoder, while heterogeneous SAR–optical features are integrated through simple feature concatenation. Based on this baseline, optical reliability modulation (ORM), heterogeneous information adaptive aggregation (HIAA), the unified semantic mapping transformer (USMT), and semantic anchor-guided optimization were gradually added. All variants were trained and evaluated under the same experimental settings described in Section 5.1.

Table 7 reports the ablation results. S1 represents the baseline model without the proposed modules, which achieves an OA of 81.35%, an F1-score of 75.92%, and an mIoU of 66.58%. After introducing ORM in S2, the mIoU increases to 68.02%, indicating that reliability-aware modulation helps reduce the negative influence of cloud-degraded optical features. This result suggests that treating all optical responses equally is suboptimal under cloud-contaminated conditions, because heavily obscured regions may provide unreliable semantic cues.

When HIAA is further added in S3, the performance improves substantially, with OA, F1-score, and mIoU reaching 84.53%, 80.37%, and 70.66%, respectively. The mIoU gain from S2 to S3 is 2.64%, which is the largest improvement among the step-wise module additions. This demonstrates that heterogeneous high-order interaction plays a central role in SAR–optical feature aggregation. Compared with simple concatenation, HIAA better models spatial and channel dependencies between cloud-contaminated optical features and SAR structural features, thereby enhancing the discriminative representation of land cover categories.

Adding USMT in S4 further improves the mIoU to 72.42%. This indicates that directly decoding fused features into LULC maps is less effective than first organizing them in a unified semantic latent space. By introducing transformer-based semantic token mapping, USMT helps capture long-range semantic relationships and reduces the representation gap between heterogeneous fused features and LULC categories. Finally, S5 achieves the best performance after introducing semantic anchor-guided optimization, with an OA of 86.60%, an F1-score of 83.29%, and an mIoU of 73.51%. Compared with S1, the full model improves OA, F1-score, and mIoU by 5.25%, 7.37%, and 6.93%, respectively. These results confirm that ORM, HIAA, USMT, and semantic anchor-guided optimization provide complementary contributions to robust cloud-contaminated SAR–optical LULC mapping.

**Table 7**

Ablation study of the key components in CloudLULC-Net. "√" indicates that the module is included, while "–" indicates that the module is excluded. Bold values indicate the best performance, and underlined values denote the second-best.

| Setting | Key components | | | | OA (%) | F1 (%) | mIoU (%) |
|---|---|---|---|---|---|---|---|
| | ORM | HIAA | USMT | Semantic anchor | | | |
| S1 | - | - | - | - | 81.35 | 75.92 | 66.58 |
| S2 | √ | - | - | - | 82.47 | 77.34 | 68.02 |
| S3 | √ | √ | - | - | 84.53 | 80.37 | 70.66 |
| S4 | √ | √ | √ | - | <u>85.71</u> | <u>82.18</u> | <u>72.42</u> |
| S5 | √ | √ | √ | √ | **86.60** | **83.29** | **73.51** |

6.3.2 Effectiveness of spatial–channel high-order interaction

To further investigate the effectiveness of the HIAA block, we conducted an ablation study on different heterogeneous interaction strategies. Since HIAA consists of the Spatial High-Order Interaction (SHOI) module and the Channel High-Order Interaction (CHOI) module, this experiment aims to verify whether spatial high-order interaction and channel-wise recalibration both contribute to SAR–optical feature aggregation. Five fusion settings were compared, including simple concatenation, conventional cross-attention fusion, HIAA without SHOI, HIAA without CHOI, and the full HIAA block. All variants were evaluated under the same experimental protocol, while the remaining components of CloudLULC-Net were kept unchanged.

Table 8 reports the quantitative results. Simple concatenation achieves an OA of 82.18%, an F1-score of 78.21%, and an mIoU of 68.74%. Although this strategy has the lowest computational cost and the highest inference speed, it only combines SAR and optical features at the channel level and lacks explicit cross-modal interaction. As a result, the complementary relationship between cloud-contaminated optical features and SAR structural cues cannot be sufficiently explored. Replacing simple concatenation with cross-attention improves the mIoU from 68.74% to 70.15%, showing that attention-based feature interaction is beneficial for heterogeneous fusion. However, cross-attention introduces much higher computational cost, with FLOPs increasing from 15.46 G to 33.72 G and FPS decreasing from 51.63 to 25.84.

When only CHOI is retained, the model achieves an mIoU of 70.88%, which is higher than simple concatenation and cross-attention. This indicates that channel-wise recalibration can enhance discriminative spectral–structural responses by adaptively weighting heterogeneous feature channels. However, without SHOI, the model lacks explicit spatial high-order interaction and is less capable of capturing long-range spatial dependencies between SAR and optical representations. When only SHOI is retained, the performance further improves to an OA of 85.38%, an F1-score of 81.87%, and an mIoU of 72.06%. This result suggests that spatial high-order interaction plays a more dominant role in cloud-contaminated SAR–optical LULC mapping, because it enables SAR-derived spatial structures to compensate for missing or unreliable optical information in cloud-obscured regions.

**Table 8**

Comparative ablation results of different heterogeneous interaction strategies in CloudLULC-Net. "√" indicates that the corresponding module is included, while "–" denotes its absence. Bold values indicate the best performance, and underlined values denote the second-best. All experiments were conducted under the same settings, with computational metrics measured using 512×512 inputs on a single NVIDIA RTX A5000 GPU.

| Setting | Fusion strategy | Key Module | | OA (%) | F1 (%) | mIoU (%) | FLPOs (G) | Param. (M) | Speed (FPS) |
|---|---|---|---|---|---|---|---|---|---|
| | | SHOI | CHOI | | | | | | |
| S1 | Simple concatenation | - | - | 82.18 | 78.21 | 68.74 | 15.46 | 22.37 | 51.63 |
| S2 | Cross-attention | √ | - | 83.64 | 79.73 | 70.15 | 33.72 | 34.58 | 25.84 |
| S3 | w/o SHOI | - | √ | 84.26 | 80.52 | 70.88 | 17.64 | 24.32 | 46.58 |
| S4 | w/o CHOI | √ | - | <u>85.38</u> | <u>81.87</u> | <u>72.06</u> | 18.93 | 23.16 | 45.01 |
| S5 | Full HIAA | √ | √ | **86.60** | **83.29** | **73.51** | 20.51 | 25.90 | 42.27 |

The full HIAA block, which combines SHOI and CHOI, achieves the best performance, with an OA of 86.60%, an F1-score of 83.29%, and an mIoU of 73.51%. Compared with the setting without SHOI, the full HIAA improves mIoU by 2.63%, and compared with the setting without CHOI, it improves mIoU by 1.45%. These results demonstrate that SHOI and CHOI are complementary: SHOI strengthens cross-modal spatial dependency modeling, while CHOI recalibrates heterogeneous channel responses after spatial interaction. In addition, the full HIAA maintains moderate computational complexity, requiring 20.51 G FLOPs and 25.90 M parameters, while achieving 42.27 FPS. This indicates that the proposed spatial–channel high-order interaction design improves classification accuracy without introducing excessive computational burden.

6.3.3. Sensitivity analysis of HIAA depth and interaction order

To investigate the influence of progressive high-order interaction, we conducted sensitivity analysis on two key hyperparameters of the HIAA module: the number of stacked HIAA blocks $N$ and the maximum interaction order $\mathcal{O}$. Specifically, $N$ was varied from 1 to 5, and $\mathcal{O}$ was varied from 2 to 7. The corresponding OA, F1-score, and mIoU results are shown in Fig. 13.

As shown in Fig. 13, increasing $N$ improves the classification performance at the early stage. When $N$ increases from 1 to 3, the performance rises consistently across OA, F1-score, and mIoU. This indicates that a single HIAA block is insufficient to fully capture heterogeneous spatial–channel dependencies between cloud-contaminated optical features and SAR structural cues. By stacking multiple HIAA blocks, the network can progressively refine SAR–optical representations and enhance the discriminative ability of fused features for LULC mapping. However, the performance does not continue to improve when $N$ is further increased. After $N = 3$, the accuracy gain becomes saturated and even decreases slightly for $N = 4$ and $N = 5$. This phenomenon may be caused by redundant feature transformations, increased optimization difficulty, or over-smoothing of heterogeneous representations after excessive iterative interaction. Therefore, $N = 3$ provides a more suitable depth for progressive SAR–optical aggregation, achieving a favorable balance between representation capacity and training stability.

The interaction order $\mathcal{O}$ controls the complexity of spatial and channel high-order feature interaction within each HIAA block. In general, increasing $\mathcal{O}$ improves model performance because higher-order interaction can capture more complex and nonlinear relationships between optical and SAR representations. This is particularly useful under cloud-contaminated conditions, where the correspondence between optical spectral cues and SAR structural information is often indirect and spatially heterogeneous. Nevertheless, the performance gain gradually becomes smaller when $\mathcal{O}$ continues to increase. Although higher orders such as $\mathcal{O} = 6$ or $\mathcal{O} = 7$ may achieve slightly higher accuracy in some cases, they also introduce higher computational complexity and may reduce training efficiency.

Considering both accuracy and efficiency, we set $N = 3$ and $\mathcal{O} = 5$ as the default configuration of HIAA in this study. This setting provides sufficiently strong high-order interaction while avoiding unnecessary model complexity caused by excessive interaction depth. The sensitivity analysis demonstrates that both progressive HIAA stacking and high-order interaction contribute to performance improvement, but their settings should be carefully balanced to achieve stable and efficient cloud-contaminated SAR–optical LULC mapping.

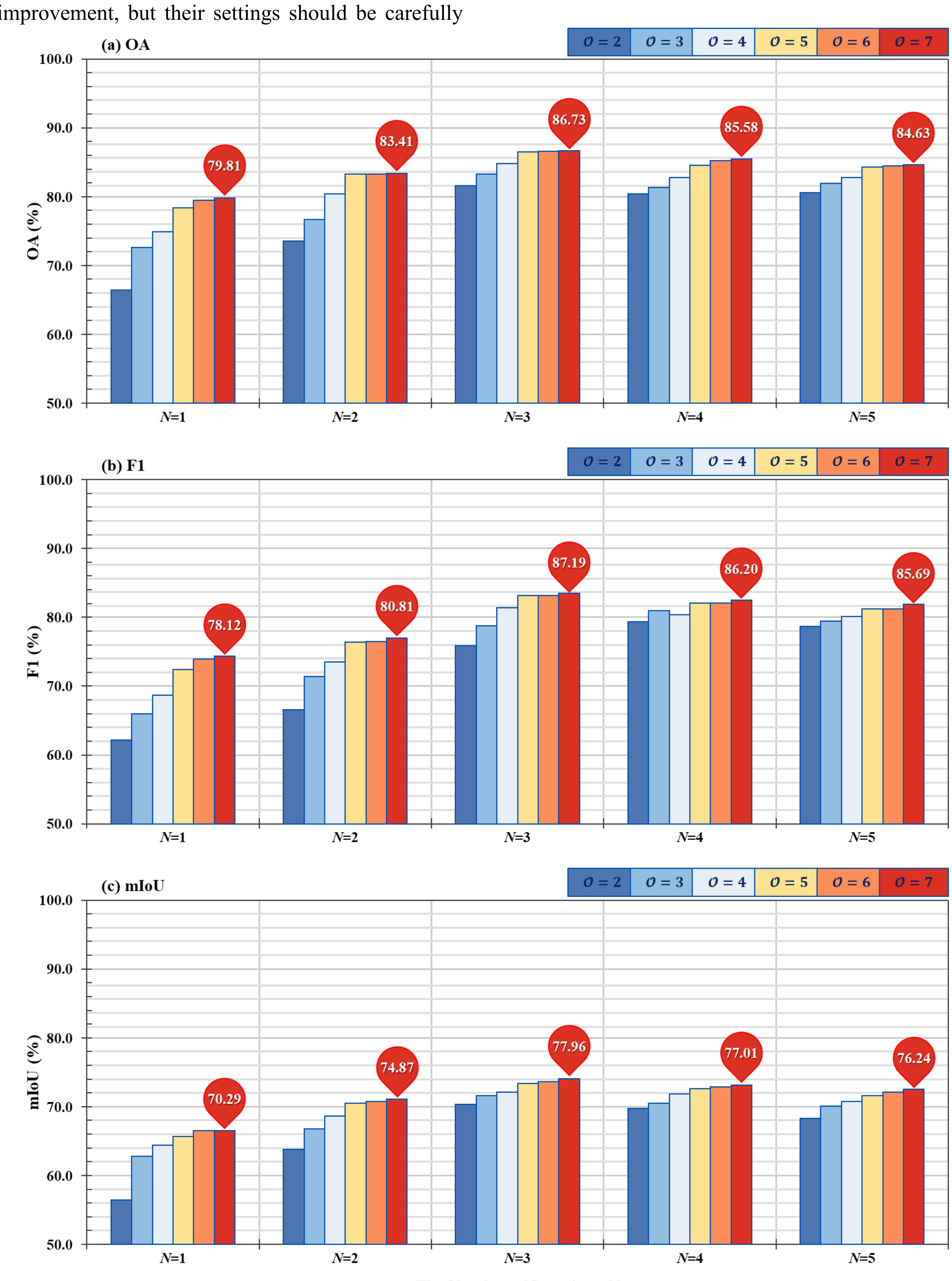

**Fig. 13.** Sensitivity analysis of CloudLULC-Net with respect to the number of stacked HIAA blocks $N$ and the maximum interaction order $O$. (a) OA, (b) F1-score, and (c) mIoU under different combinations of $N$ and $O$.

### 6.4. Benefits of LULC mapping for fine-grained spatiotemporal surface analysis

Beyond benchmark accuracy, an important advantage of CloudLULC-Net lies in its potential for fine-grained target-date LULC mapping in regions with frequent cloud contamination and rapid seasonal surface changes. Existing global LULC products are valuable for long-term and large-area monitoring, but many of them are generated at annual or product-level temporal intervals. This may limit their ability to characterize short-term land surface dynamics, especially in regions where valid optical observations are frequently interrupted by clouds.

To illustrate this practical value, we selected the Poyang Lake Basin in Jiangxi Province, China, as a case study area. The geographical location of this region is shown in Fig. 2. The study area covers approximately $109.8 \times 109.8$ km$^2$ and is located in a subtropical monsoon climate zone, where seasonal hydrological fluctuations and frequent cloud contamination jointly increase the difficulty of LULC mapping. Four target scenes in 2024 were selected to represent different seasonal stages, including January, April, July, and October. For each target scene, CloudLULC-Net was applied using the cloud-contaminated Sentinel-2 observation and the temporally adjacent Sentinel-1 SAR image to generate seasonal LULC maps.

Google Dynamic World was included as a representative near-

real-time optical-based LULC product for comparison. Dynamic World provides 10 m near-real-time land cover predictions from Sentinel-2 observations, making it highly valuable for monitoring rapidly changing land surfaces. However, its effective availability is still constrained by optical observation conditions. According to the Dynamic World data description, its predictions are generated only from Sentinel-2 L1C observations satisfying a low-cloud screening criterion, i.e., images with no more than 35% cloudy pixels, while cloud and cloud-shadow pixels are further masked during product generation. Therefore, in persistently cloudy regions or seasons with frequent high-cloud observations, Dynamic World may provide temporally discontinuous or locally incomplete LULC results, even though it is designed for near-real-time monitoring. This limitation highlights the need for a cloud-tolerant target-date mapping framework that can directly exploit available cloud-contaminated optical observations together with SAR data.

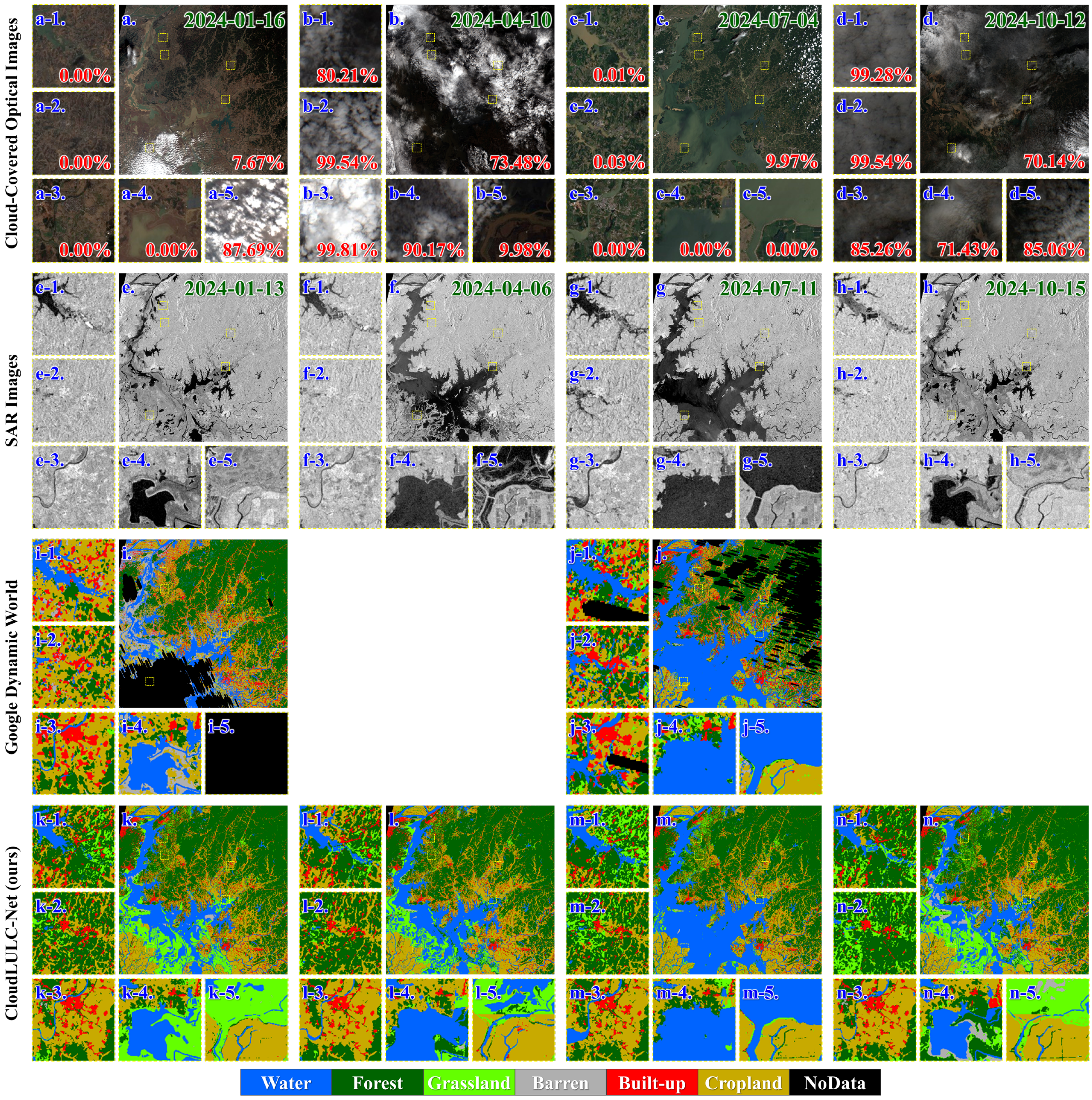

**Fig. 14.** Seasonal LULC mapping results in the Poyang Lake Basin generated by CloudLULC-Net and Google Dynamic World. The first two rows show cloud-contaminated optical images and corresponding SAR images acquired in different seasons. The following rows present LULC maps generated by Google Dynamic World and CloudLULC-Net. Red numerals indicate cloud-cover percentages, and green numerals denote image acquisition dates.

As shown in Fig. 14, the optical observations of the Poyang Lake Basin exhibit substantial seasonal differences in both land surface conditions and cloud contamination. In January and July, relatively low-cloud observations are available in most local areas, whereas the April and October scenes contain extensive cloud-covered regions. The corresponding SAR images preserve structural information related to water boundaries, terrain texture, and built-up patterns under cloudy conditions. By integrating cloud-contaminated Sentinel-2 imagery with temporally adjacent Sentinel-1 SAR observations, CloudLULC-Net generates spatially continuous seasonal LULC maps for all four target dates. The seasonal mapping results reveal clear land surface dynamics in the Poyang Lake Basin. The spatial extent of water bodies changes noticeably across seasons, reflecting the hydrological fluctuation of the lake basin. Vegetation and cropland patterns also vary with phenological stages, while built-up areas remain relatively stable. Compared with Google Dynamic World, which may be unavailable or incomplete when the corresponding Sentinel-2 observations do not satisfy low-cloud conditions, CloudLULC-Net provides more continuous LULC information across the selected seasonal observations. This capability is particularly important for regions where land surface conditions change rapidly and where waiting for fully cloud-free optical observations may cause temporal delays.

This case study demonstrates that CloudLULC-Net can complement existing global LULC products by improving the temporal availability of target-date land cover information in cloud-prone regions. Rather than relying solely on low-cloud optical observations, the proposed framework directly uses cloud-contaminated optical imagery and temporally adjacent SAR data, enabling fine-grained seasonal LULC mapping for applications such as hydrological monitoring, agricultural assessment, and regional land surface change analysis.

### 6.5. Limitations and prospects

Although CloudLULC-Net shows promising performance for cloud-contaminated SAR–optical LULC mapping, several limitations remain. First, the current framework is mainly developed and evaluated using Sentinel-1 C-band SAR and Sentinel-2 multispectral observations. Its generalization to other SAR frequencies, optical sensors, imaging geometries, and spatial resolutions still requires further validation. Second, the method relies on temporally adjacent SAR and optical observations. While this setting is generally suitable for LULC mapping, temporal mismatch may introduce uncertainty in regions with rapid surface changes, such as floodplains, agricultural transition zones, and urban expansion areas.

Future work will focus on improving the scalability and robustness of cloud-contaminated LULC mapping. Incorporating longer SAR–optical time series may help model temporal consistency and reduce the influence of acquisition mismatch. In addition, semi-supervised learning, domain adaptation, and uncertainty estimation could be explored to reduce dependence on dense annotations and improve transferability to unseen regions or sensor platforms. These extensions may further enhance the applicability of CloudLULC-Net for operational, fine-grained, and near-real-time LULC mapping in cloud-prone regions.

## 7. Conclusions

This study investigated the problem of near-real-time LULC mapping under cloud-contaminated optical observation conditions and proposed CloudLULC-Net, a heterogeneous SAR–optical fusion framework designed to directly predict LULC maps from cloud-contaminated Sentinel-2 imagery and temporally adjacent Sentinel-1 SAR observations. To support this task, we constructed CloudLULC-Set, a large-scale benchmark dataset consisting of curated SAR–optical–label triplets with pixel-level LULC annotations. Methodologically, CloudLULC-Net integrates optical reliability modulation, heterogeneous information adaptive aggregation, unified semantic mapping, and semantic anchor-guided optimization to suppress unreliable optical responses, exploit SAR-derived structural cues, and align fused representations with LULC-oriented semantic space. Extensive experiments demonstrate that CloudLULC-Net outperforms representative heterogeneous reconstruction-first and end-to-end SAR–optical mapping methods in terms of both classification accuracy and computational efficiency. Comparative validation with existing global LULC products further confirms its practical value for target-date land cover mapping in cloud-prone regions. Additional analyses show that the proposed framework maintains stronger robustness under increasing cloud coverage, benefits substantially from SAR observations, and provides useful support for fine-grained seasonal surface analysis. Future work will focus on improving cross-sensor generalization, incorporating longer SAR–optical time series, and enhancing model transferability under broader geographical and observation conditions.

## CRediT authorship contribution statement

**Jiangong Xu:** Writing - original draft, Software, Methodology, Investigation, Data curation, Conceptualization. **Weibao Xue:** Writing - review & editing, Investigation, Methodology, Validation. **Xiaoyu Yu:** Writing - review & editing, Methodology, Investigation. **Jun Pan:** Validation, Supervision, Project administration, Funding acquisition, Formal analysis, Conceptualization. **Xinlian Liang:** Writing-review & editing, Supervision. **Mi Wang:** Writing - review & editing, Supervision.

## Declaration of competing interest

The authors declare that they have no known competing financial interests or personal relationships that could have appeared to influence the work reported in this paper.

## Data availability

The CloudLULC-Set dataset employed in this research is openly accessible via its official GitHub repository: https://github.com/RSIIPAC/CloudLULC.

## Acknowledgments

The authors gratefully acknowledge the European Space Agency (ESA) for providing high-quality satellite imagery essential to this study. This work was supported by the National Natural Science Foundation of China under Grant 62371352, and 62425102.